\documentclass{article}

\usepackage{worldts-preprint,times}

\usepackage{amsmath,amsfonts,bm}

\def\eqref#1{equation~\ref{#1}}

\def\1{\bm{1}}

\DeclareMathAlphabet{\mathsfit}{\encodingdefault}{\sfdefault}{m}{sl}
\SetMathAlphabet{\mathsfit}{bold}{\encodingdefault}{\sfdefault}{bx}{n}

\usepackage{hyperref}
\usepackage{url}
\usepackage{graphicx}
\usepackage{wrapfig}
\usepackage{booktabs}
\usepackage{xcolor}
\definecolor{bestred}{HTML}{C00000}
\definecolor{secondblue}{HTML}{0057B8}
\DeclareRobustCommand{\bestresult}[1]{\textcolor{bestred}{\textbf{#1}}}
\DeclareRobustCommand{\secondresult}[1]{\textcolor{secondblue}{\underline{#1}}}
\usepackage{amsmath}
\usepackage{amssymb}
\usepackage{multirow}
\usepackage{algorithm}
\usepackage{algorithmic}

\title{WorldTS: World Modeling for Multimodal Covariate-aware Time Series Forecasting}

\author{
Yuhan Zhu\textsuperscript{1}, Xiangfei Qiu\textsuperscript{1}, Hanyin Cheng\textsuperscript{1}, Wongmeng Shen\textsuperscript{1}\\
\textbf{Chenjuan Guo\textsuperscript{1}, Bin Yang\textsuperscript{1}, Jilin Hu\textsuperscript{1}, Christian S. Jensen\textsuperscript{2}}\\
\normalfont\textsuperscript{1}East China Normal University\\
\normalfont\textsuperscript{2}Aalborg University
}

\worldtsfinalcopy

\begin{document}
\maketitle
\fancyhead{}
\fancyhead[L]{Preprint}
\renewcommand{\headrulewidth}{0.4pt}


\begin{abstract}

Time series forecasting is typically framed as learning a direct mapping from historical to future observations in the observation space. However, sequences of observations generally provide only a partial view of the dynamics of the underlying system, with future observations being shaped by latent dynamics. Recent latent-space forecasting methods thus achieve improved performance by predicting future observations from latent-space representations of historical observations rather than directly forecasting future observations in the observation space. Next, while future observations are also shaped by external factors, how to incorporate external, often multimodal, information into forecasting, so that it can shape latent-state formation and evolution directly, remains underexplored. We propose WorldTS, a world-modeling based forecasting framework that integrates multimodal covariates directly into the forecasting to further improve forecasting performance. Specifically, WorldTS employs a two-stage training strategy. First, it learns forecasting-relevant latent state dynamics conditioned on multimodal covariates, yielding encoded future states. Next, the learned state dynamics are frozen, and an observation decoder is trained to map the predicted future states back to future observations. Extensive experiments on 21 real-world datasets offer insight into WorldTS and its effectiveness.

\end{abstract}


\section{Introduction}

Time series forecasting plays a fundamental role in a wide range of applications, including
energy~\citep{alvarez2010energy,cheng2026metagnsdformer},
transportation~\citep{yu2025merlin,qiu2024tfb},
climate~\citep{wu2023weathergnn,liu2026astgi},
finance~\citep{wu2025k2vae,wu2026timeart},
industrial monitoring~\citep{wu2025catch,qiu2025tab},
and public governance~\citep{zheng2015forecasting,xu2026most}.
Traditional time series forecasting learns a direct mapping from historical observations to future observations in the observation space~\citep{qiu2025DBLoss,wu2025srsnet,liu2026rethinking}.
However, observed observations are only external manifestations of the underlying states of dynamic systems, rather than the states themselves~\citep{ghugare2022simplifying,kaiser2020model}.
For many systems, identical or similar observations may correspond to different underlying states, while future evolution largely depends on these states and their dynamics.
Directly forecasting future observations from historical observations therefore entangles state evolution with its manifestation in the observation space.
As a result, models may focus more on fitting numerical relationships in the observation space than on explicitly modeling the underlying system states and their dynamics.

This reasoning motivates a more natural alternative: explicitly modeling future states in a latent space, shifting time series forecasting towards modeling system states and their dynamics.
Specifically, a model first extracts state representations from historical observations, predicts their future evolution in the latent space, and then decodes the predicted states back into future observations.
Compared with directly forecasting future observations, this formulation treats future states as prediction targets, placing emphasis on latent-state evolution and partially decoupling state dynamics from observation generation.
Modeling system evolution through future-state prediction in a latent space has become an important paradigm in world modeling~\citep{assran2023ijepa,maes2026leworldmodel}. Similarly, in time series forecasting, mapping historical observations into a latent state space and modeling future states in that space can better capture system states and dynamics, thereby improving forecasting performance~\citep{yang2026latentsf}.

\begin{figure}[t]
    \centering
    \includegraphics[width=\textwidth]{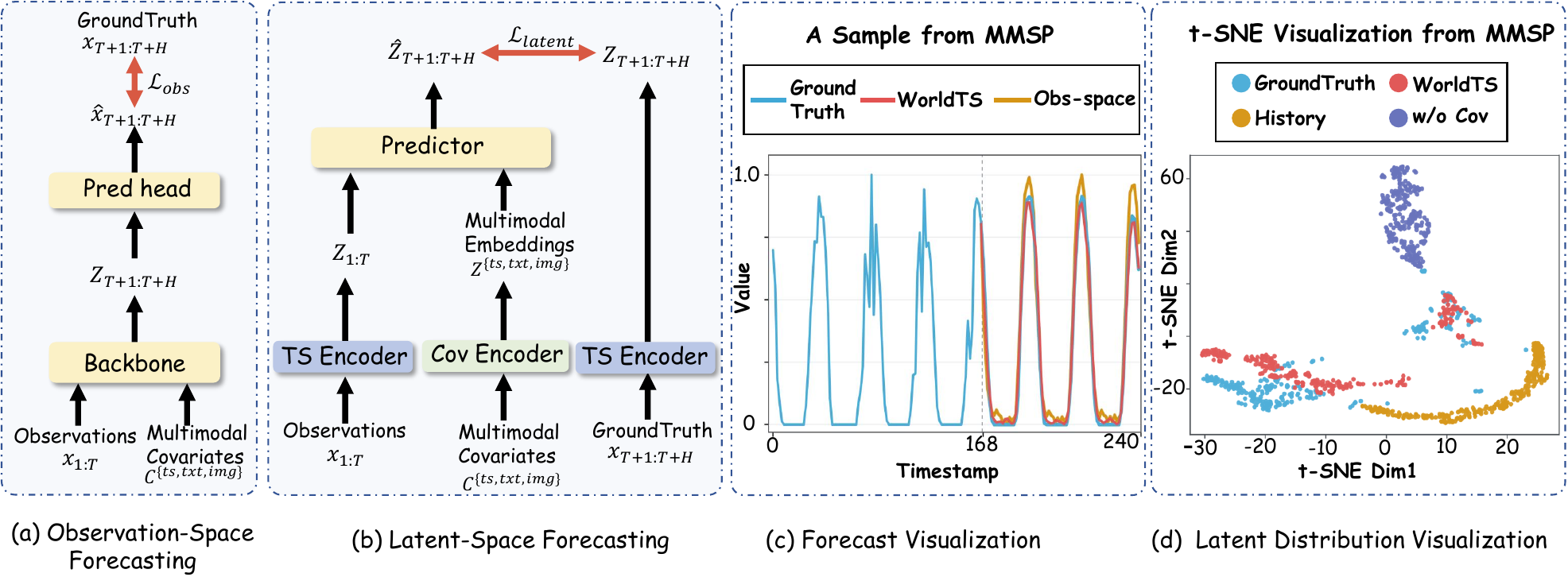}
    \caption{
    Comparison of observation-space and latent-space forecasting.
    (a) Observation-space forecasting supervised by ground-truth future observations.
    (b) Latent-space forecasting supervised by states encoded from the ground-truth future observations.
    (c) Forecasts on the Multimodal Solar Power (MMSP) dataset by WorldTS and its observation-space variant (Obs-space).
    (d) Latent distributions on MMSP, comparing predicted future states with states encoded from the ground-truth future observations.    }
    \label{fig:intro}
\end{figure}

However, in real-world time series forecasting future states are rarely determined by historical states alone; they are also shaped by external factors.
Historical observations describe past evolution, whereas external factors in the form of covariates provide additional predictive information relevant to future evolution.
For example, electricity-load dynamics depend not only on historical consumption patterns but also on external weather conditions, while photovoltaic power generation is affected by cloud cover~\citep{ma2024fusionsf,wang2026vot}.
Forecasting future states solely from the historical observations may overlook important factors governing system evolution, and incorporating evolution-relevant covariates can provide richer information for modeling future system states.

Covariates often span multiple modalities. Different modalities can provide complementary information that is difficult to infer from historical observations, thus offering additional cues for predicting future system evolution~\citep{liu2024timemmd,ma2024fusionsf}.
Latent-space modeling provides a natural way to integrate multimodal information: diverse modalities can be encoded into compact representations that can then contribute to future-state prediction~\citep{han2026unica,li2026tats}.
Consequently, multimodal covariates can enrich state modeling beyond historical data, and their latent representations can be incorporated naturally into future-state prediction.

Motivated by these observations, we propose \textbf{WorldTS}, a world-modeling framework for multimodal, covariate-aware time series forecasting.
As illustrated in Figure~\ref{fig:intro}(a), traditional covariate-aware forecasting predicts future observations directly and is optimized in the observation space.
WorldTS instead performs prediction in a latent state space.
As shown in Figure~\ref{fig:intro}(b), 
In Stage 1, the shared TS Encoder maps both the historical observations and the ground-truth future observations into the latent state space, while modality-specific encoders (Cov Encoder) map multimodal covariates into their corresponding representations. The historical states and covariate representations are then used to predict future states, which are directly supervised by the states encoded from the ground-truth future observations. In Stage~2, we freeze the encoders and predictor and train a decoder to map the predicted future states back to the observation space. Figures~\ref{fig:intro}(c)--(d) further illustrate how this design affects both the predicted states and the final forecasts.
Figure~\ref{fig:intro}(d) compares the latent distributions of the predicted future states and the states encoded from the ground-truth future observations.
With covariates, the predicted future states exhibit a distribution more consistent with that of the encoded future states, whereas removing covariates results in a more pronounced distribution discrepancy.
Figure~\ref{fig:intro}(c) shows the corresponding forecasting results in the observation space.
Here, Obs-space denotes a one-stage variant of WorldTS trained end-to-end using only an observation-space loss, without direct supervision on future states.
Compared with Obs-space, WorldTS produces a forecast trajectory that more closely matches the ground-truth future sequence.

In summary, our main contributions are as follows:
\begin{itemize}
    \item We formulate multimodal covariate-aware time series forecasting from a latent-space future-state prediction perspective, where multimodal covariates condition future target-state prediction rather than being used only for observation-space forecasting.

    \item We propose \textbf{WorldTS}, a world-modeling framework featuring direct supervision from states encoded from future observations and a two-stage training strategy.
    WorldTS first learns covariate-conditioned state dynamics and then decodes predicted states into future observations.

    \item We report on extensive experiments on 21 real-world datasets, covering numerical-, text-, and image-covariate forecasting settings.
    Comprehensive comparisons and ablation studies characterize the effectiveness of WorldTS.
\end{itemize}

\section{Related Work}

\subsection{Latent-Space Predictive Modeling}

Latent-space prediction is central to world modeling, where observations are
encoded as latent states whose future evolution is predicted.
DreamerV3~\citep{hafner2023dreamerv3} learns latent state transitions
from observation and action histories, supporting decision-making
through imagined state trajectories.
JEPA-based methods use encoded observations as explicit prediction targets.
V-JEPA 2-AC~\citep{assran2025vjepa2} uses a pretrained video encoder to
obtain state representations and predicts future frame representations
from historical states and actions.
LeWorldModel~\citep{maes2026leworldmodel} learns latent states and their
transitions directly from images, using encoded future observations to
supervise action-conditioned state prediction.

Latent-space predictive modeling has also been explored for time series forecasting
through representation alignment and future-state prediction.
TimeAlign~\citep{hu2026timealign} aligns representations learned from historical inputs with those encoded from true future targets to guide representation learning.
LatentTSF~\citep{yang2026latentsf} maps historical and future sequences
into a shared latent space, predicts future states within this space,
and decodes them back into observations.
These methods primarily encode and predict latent states from the target
series itself.
In contrast, WorldTS further incorporates multimodal covariates into future-state prediction,
enabling the use of additional information beyond historical observations.

\subsection{Covariate-aware Forecasting}

Covariate-aware forecasting leverages external information to improve prediction. For numerical covariates, existing methods focus mainly on modeling interactions between endogenous and exogenous variables or on exploiting known future information. TimeXer~\citep{wang2024timexer} and CrossLinear~\citep{zhou2025crosslinear} capture dependencies between target and exogenous series, while TFT~\citep{lim2021tft} and TiDE~\citep{das2023tide} explicitly accommodate known future covariates. DAG~\citep{qiu2026dag} further models cross-variable dependencies along both temporal and channel dimensions, GCGNet~\citep{li2026gcgnet} introduces graph-structured consistency, and KITE~\citep{cheng2026kite} incorporates external information into probabilistic forecasting.

Recent studies extend covariate-aware forecasting beyond numerical variables to cover heterogeneous modalities.
TaTS~\citep{li2026tats} and VoT~\citep{wang2026vot} exploit textual information through temporal narratives and event-driven reasoning, respectively, while FusionSF~\citep{ma2024fusionsf} incorporates heterogeneous sources such as satellite imagery.
UniCA~\citep{han2026unica} provides a unified approach for adapting time-series foundation models to heterogeneous covariates.
While these methods differ considerably, they primarily incorporate covariates as additional information for observation-space forecasting.
In contrast, WorldTS considers latent-space future-state prediction: multimodal covariates are encoded into latent representations and are used to condition future-state prediction, while states encoded from ground-truth future observations provide direct supervision for the predicted states. This latent-space formulation provides a natural way to integrate heterogeneous multimodal information into future-state modeling.

\section{Preliminaries}
\subsection{Problem Formulation}
\paragraph{Multimodal Covariate-aware Forecasting.}
Given a historical time series
$x_{1:T}\in\mathbb{R}^{T\times C}$ with $C$ variables, the goal is to
forecast $x_{T+1:T+H}\in\mathbb{R}^{H\times C}$ over the next $H$ time steps
using multimodal covariates: 
\begin{equation}
    \mathcal{C}
    =
    \left\{
    c^m_{1:\tau_m}
    \right\}_{m\in\mathcal{M}},
    \qquad
    \mathcal{M}
    =
    \{\mathrm{ts},\mathrm{txt},\mathrm{img}\},
    \label{eq:covariates}
\end{equation}
where $\tau_m$ denotes the temporal availability of modality $m$.\footnote{
Covariates may be future-unknown ($\tau_m=T$) or future-known
($\tau_m=T+H$).}
The forecasting task is formulated as
\begin{equation}
    \widehat{x}_{T+1:T+H}
    =
    F_{\theta}\!\left(
    x_{1:T},
    \mathcal{C}
    \right),
    \label{eq:forecasting}
\end{equation}
where $F_{\theta}$ denotes a forecasting model parameterized by $\theta$.

\paragraph{Latent-State Forecasting.}
WorldTS further formulates the above task as forecasting future states
in a learned latent space. The historical observations and the covariates
from each modality are first mapped to their corresponding representations:
\begin{equation}
    \begin{aligned}
        Z_{1:T}
        &= E_{\theta}(x_{1:T})
        \in \mathbb{R}^{T \times D},
        &
        Z^m_{1:\tau_m}
        &= E_{\theta_m}^{m}\!\left(c^m_{1:\tau_m}\right),
        \quad m\in\mathcal M
    \end{aligned}
    \label{eq:historical_target_state}
\end{equation}
Here, $E_{\theta}$ is the state encoder,
$E_{\theta_m}^{m}$ is the encoder for modality $m$, and $D$ is the latent-state
dimensionality. Conditioned on the historical states and encoded multimodal
covariates, the predictor $P_{\psi}$ predicts future states, which
are subsequently mapped back to the observation space by the decoder
$D_{\phi}$:
\begin{equation}
    \widehat{Z}_{T+1:T+H}
    =
    P_{\psi}
    \left(
        Z_{1:T},
        \left\{Z^m_{1:\tau_m}\right\}_{m\in\mathcal M}
    \right),
    \qquad
    \widehat{x}_{T+1:T+H}
    =
    D_{\phi}
    \left(
        \widehat{Z}_{T+1:T+H}
    \right)
    \label{eq:latent_forecasting}
\end{equation}

During training, the same state encoder encodes the ground-truth future
observations to obtain
\begin{equation}
    Z_{T+1:T+H}
    =
    E_{\theta}(x_{T+1:T+H})
    \in
    \mathbb{R}^{H \times D}
    \label{eq:future_state}
\end{equation}
which provides direct supervision for
$\widehat{Z}_{T+1:T+H}$.

\section{WorldTS}
\label{sec:method}

\begin{figure*}[t]
    \centering
    \includegraphics[width=0.9\textwidth]{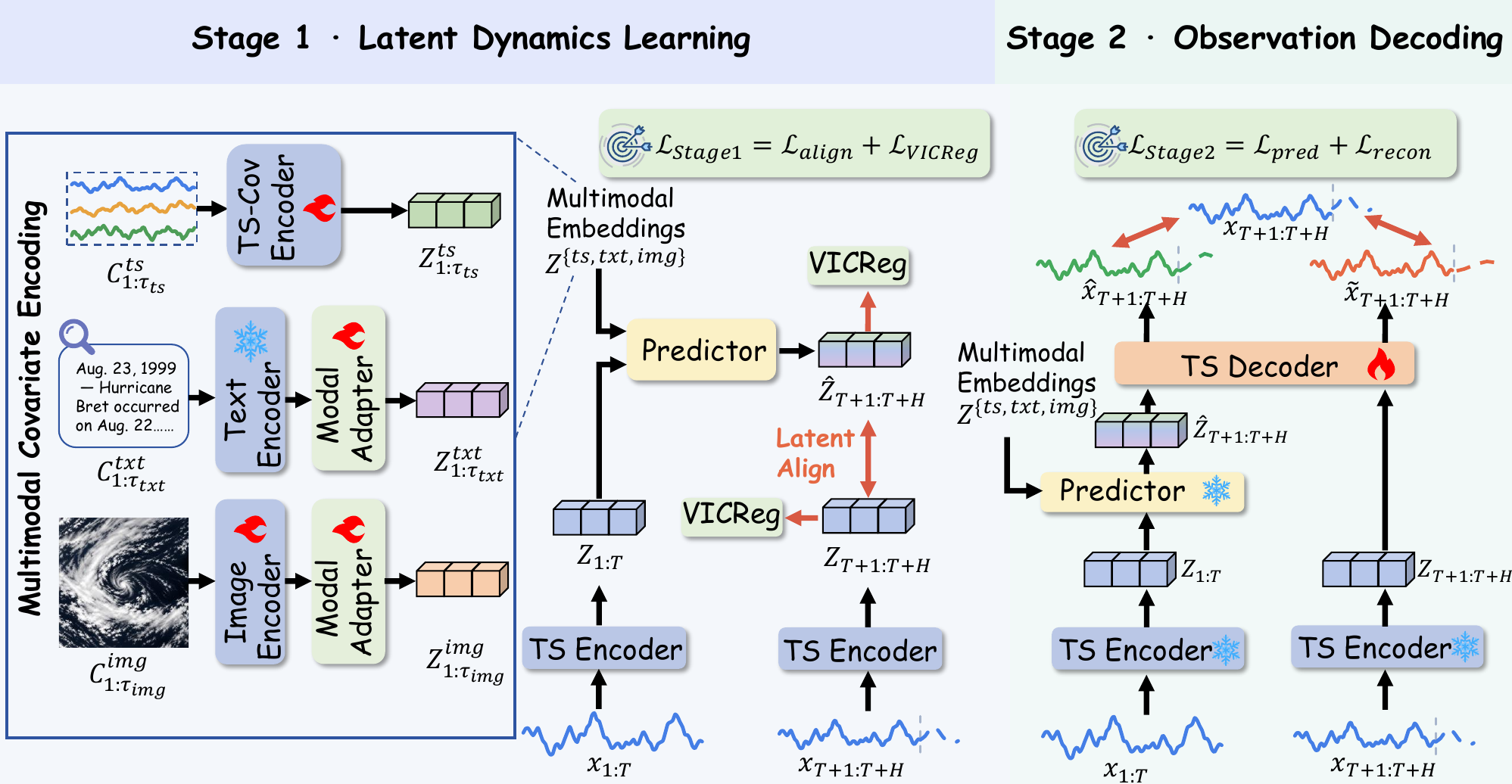}
    \caption{Overview of WorldTS. TS Encoder/Decoder denote the time-series
    encoder/decoder; Predictor denotes the covariate-conditioned state predictor.
    Cov Encoder and Modal Adapter are the numerical-covariate encoder and
    modality-specific MLP projector. Stage~1 (Latent Dynamics
    Learning) learns future-state prediction conditioned on multimodal covariates.
    Stage~2 (Observation Decoding) freezes the encoders and predictor and trains
    the decoder to recover future observations.}
    \label{fig:worldts_framework}
\end{figure*}

\subsection{Framework Overview}
\label{sec:framework_overview}

As shown in Figure~\ref{fig:worldts_framework}, WorldTS first learns state dynamics conditioned on multimodal covariates and then decodes the future states into observations. Specifically, its first stage, Latent Dynamics Learning, learns latent dynamics conditioned on multimodal covariates. Historical observations and multimodal covariates are encoded through separate branches into historical states and covariate representations. The predictor combines these representations to predict future states. Meanwhile, the shared TS encoder encodes the ground-truth future observations to directly supervise state prediction. State-matching losses and anti-collapse regularization jointly optimize the learnable encoding modules and predictor. In its second stage, Observation Decoding, WorldTS learns to decode future states into observations. The encoders and predictor are frozen, and only the decoder is trained. Predicted future states and states encoded from the ground-truth future observations pass through the same decoder. Both outputs are supervised by ground-truth future observations, yielding prediction and reconstruction losses, respectively. During inference, WorldTS predicts future states using only the historical observations and available covariates, then decodes them into future observations.

We next elaborate on the key modules introduced above, including the TS Encoder and TS Decoder (Section~\ref{sec:ts_encoder_decoder}), multimodal covariate encoders (Section~\ref{sec:multimodal_covariate_encoding}), and covariate-conditioned state predictor (Predictor, Section~\ref{sec:covariate_conditioned_state_predictor}), followed by the two-stage training objectives (Section~\ref{sec:two_stage_learning}).

\subsection{TS Encoder and TS Decoder}
\label{sec:ts_encoder_decoder}

In Stage 1, the shared TS Encoder encodes the historical observations and the ground-truth future observations into a common latent space. In Stage~2, the TS Decoder learns to
map future states back to observations. Both use local patches to capture
temporal context~\citep{ekambaram2023tsmixer}; their architectures are illustrated
in Appendix~\ref{app:implementation_details} (Figure~\ref{fig:state_encoder_decoder}).

We apply $\operatorname{CausalPatching}_{S_e}$ to the observations
$x_{1:L}\in\mathbb{R}^{L\times C}$, extracting $L$ overlapping
length-$S_e$ patches with stride one and replicate padding at the beginning. The patch ending at time $t$ contains only observations up to $t$,
thereby preserving causality. Each patch is flattened and mapped through a
shared linear projection into the $D$-dimensional state space:
\begin{equation}
    U^x
    = \operatorname{Linear}_e\!\left(
        \operatorname{CausalPatching}_{S_e}(x_{1:L})
    \right)
    \in \mathbb{R}^{L\times D}
    \label{eq:state_encoder_projection}
\end{equation}

A residual two-layer feed-forward network with GELU activation then refines
the projected patches:
\begin{equation}
    Z_{1:L}
    = E_{\theta}(x_{1:L})
    = U^x + \operatorname{FFN}_e(U^x)
    \in \mathbb{R}^{L\times D}
    \label{eq:state_encoder_sequence}
\end{equation}

The same TS Encoder maps the historical observations and the ground-truth future observations into a shared
state space, with the last $S_e-1$ historical observations providing context
for the first future patches. This produces the historical states
$Z_{1:T}$ and future states
$Z_{T+1:T+H}$, which supervise the predicted states
$\widehat{Z}_{T+1:T+H}$ during training.

For decoding with patch length $S_d$, we concatenate the last $S_d-1$
historical states with the predicted future states:
\begin{equation}
    \overline{Z}
    = \left[
        Z_{T-S_d+2:T};
        \widehat{Z}_{T+1:T+H}
    \right]
    \in \mathbb{R}^{(S_d-1+H)\times D}
    \label{eq:state_decoder_context}
\end{equation}

The decoder extracts $H$ consecutive length-$S_d$ state patches from
$\overline{Z}$, one per forecast step, and projects each patch into a
$D$-dimensional decoded representation:
\begin{equation}
    U^z
    = \operatorname{Linear}_d\!\left(
        \operatorname{Patching}_{S_d}(\overline{Z})
    \right),
    \qquad
    U^z \in \mathbb{R}^{H\times D}
    \label{eq:state_decoder_projection}
\end{equation}

A residual two-layer feed-forward network with GELU activation refines these
representations:
\begin{equation}
    V^z
    = U^z + \operatorname{FFN}_d(U^z)
    \label{eq:state_decoder_ffn}
\end{equation}

The output projection finally maps the refined representations from the state
dimensionality $D$ to the observation dimensionality $C$:
\begin{equation}
    \widehat{x}_{T+1:T+H}
    = D_{\phi}\!\left(
        \widehat{Z}_{T+1:T+H}
    \right)
    = \operatorname{Linear}_o(V^z)
    \in \mathbb{R}^{H\times C}
    \label{eq:state_decoder_output}
\end{equation}

\subsection{Multimodal Covariate Encoding}
\label{sec:multimodal_covariate_encoding}

In parallel with the TS Encoder, modality-specific encoders map covariates
into representations for the Predictor. The numerical-covariate encoder
(Cov Encoder) follows the TS Encoder architecture with separate parameters:
\begin{equation}
    \begin{aligned}
        U^{\mathrm{ts}}
        &= \operatorname{Linear}_{\mathrm{ts}}\!\left(
            \operatorname{CausalPatching}_{S_e}
            \!\left(c^{\mathrm{ts}}_{1:\tau_{\mathrm{ts}}}\right)
        \right) \\
        Z^{\mathrm{ts}}_{1:\tau_{\mathrm{ts}}}
        &= E_{\theta_{\mathrm{ts}}}^{\mathrm{ts}}
            \!\left(c^{\mathrm{ts}}_{1:\tau_{\mathrm{ts}}}\right)
         = U^{\mathrm{ts}}
           + \operatorname{FFN}_{\mathrm{ts}}(U^{\mathrm{ts}})
    \end{aligned}
    \label{eq:ts_covariate_encoder}
\end{equation}
The text branch follows VoT~\citep{wang2026vot} and extracts textual features
using the input embedding layer of a frozen GPT-2 model, whereas the image
branch follows UniCA~\citep{han2026unica} and employs a lightweight CNN to
extract visual features. Each branch then maps the extracted features through
a lightweight MLP projector (Modal Adapter) to obtain $Z^{\mathrm{txt}}$ or
$Z^{\mathrm{img}}$. Architectural and implementation details are provided in
the appendix.

\subsection{Covariate-Conditioned State Predictor}
\label{sec:covariate_conditioned_state_predictor}
\begin{wrapfigure}{r}{0.50\textwidth}
    \centering
    \includegraphics[width=\linewidth]{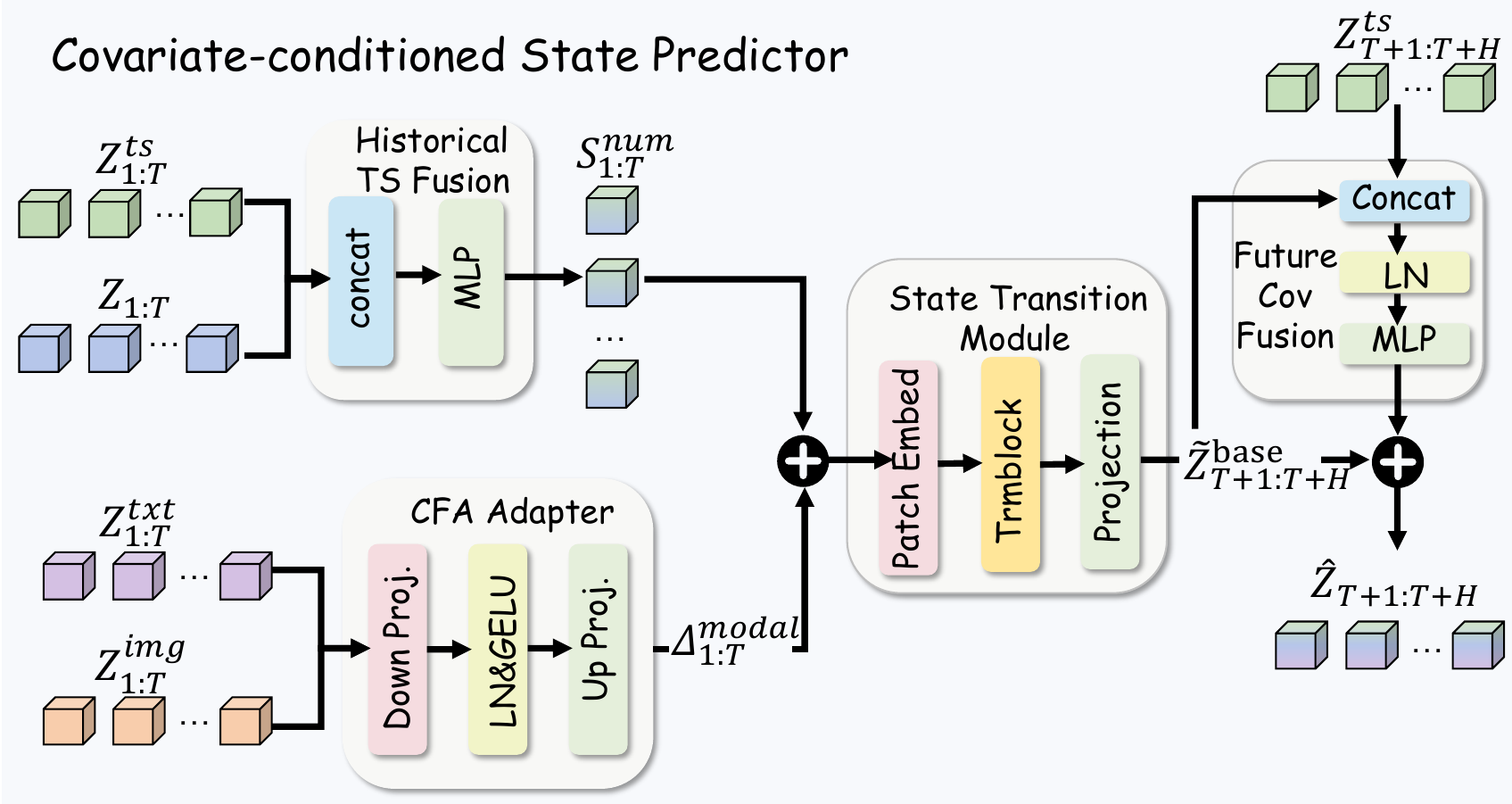}
    \caption{Covariate-conditioned state predictor. Historical covariates are integrated into the historical states through direct fusion or CFA, while future covariates further refine the predicted states.}
    \label{fig:multimodal_state_predictor}
\end{wrapfigure}

As shown in Figure~\ref{fig:multimodal_state_predictor}, the Predictor $P_\psi$
fuses outputs from the preceding encoders to predict base future states,
then refines them with future covariates when available.

Specifically, the historical states
$Z_{1:T}$ and encoded historical numerical covariates
$Z^{\mathrm{ts}}_{1:T}$ are concatenated along the feature dimension
and mapped into fused historical states
$S^{\mathrm{num}}_{1:T}\in\mathbb{R}^{T\times D}$:
\begin{equation}
    S^{\mathrm{num}}_{1:T}
    = \operatorname{MLP}_{h}\!\left(
        \operatorname{Concat}\!\left[
            Z_{1:T};
            Z^{\mathrm{ts}}_{1:T}
        \right]
    \right)
    \label{eq:historical_ts_fusion}
\end{equation}
where $\operatorname{MLP}_{h}$ uses layer normalization and GELU activation
to produce the fused historical states $S^{\mathrm{num}}_{1:T}$.
Without numerical covariates, we set
$S^{\mathrm{num}}_{1:T}=Z_{1:T}$.

The encoded text and image representations
$Z^{\mathrm{modal}}_{1:T}
\in \{Z^{\mathrm{txt}}_{1:T}, Z^{\mathrm{img}}_{1:T}\}$ are aligned with the
historical timeline. To filter information irrelevant to the state dynamics,
the Controlled Fusion Adapter (CFA)~\citep{lee2026cfa} uses a low-rank bottleneck
to map these representations into residuals in the state space:
\begin{equation}
    \Delta^{\mathrm{modal}}_{1:T}
    =
      \operatorname{Dropout}\!\left(
        \sigma\!\left(
          \operatorname{LN}\!\left(
            Z^{\mathrm{modal}}_{1:T} W_{\mathrm{down}}
          \right)
        \right)
      \right) W_{\mathrm{up}}
    \in \mathbb{R}^{T\times D},
    \label{eq:cfa_modal_residual}
\end{equation}
where $W_{\mathrm{down}}$ projects the modality representation into a
low-dimensional bottleneck, $W_{\mathrm{up}}$ maps it to the
$D$-dimensional state space, and $\sigma$ denotes GELU activation.
The resulting $\Delta^{\mathrm{modal}}_{1:T}$ is added to
$S^{\mathrm{num}}_{1:T}$, introducing text or image information while
retaining the historical state as the primary predictive signal.

The fused historical states are then processed by the state-transition
module $P_{\psi}^{\mathrm{trans}}$, which follows a patch-based Transformer
design~\citep{nie2023patchtst} to predict the base future-state trajectory:
\begin{equation}
    \widetilde{Z}^{\mathrm{base}}_{T+1:T+H}
    = P_{\psi}^{\mathrm{trans}}\!\left(
        S^{\mathrm{num}}_{1:T}
        + \Delta^{\mathrm{modal}}_{1:T}
    \right)
    \in \mathbb{R}^{H\times D}
    \label{eq:base_state_prediction}
\end{equation}
When future covariates are available, their representations are combined
with the base future states to compute a state residual:
\begin{equation}
    \Delta^{\mathrm{future}}_{T+1:T+H}
    = \operatorname{MLP}_{f}\!\left(
        \operatorname{LN}\!\left(
          \operatorname{Concat}\!\left[
            \widetilde{Z}^{\mathrm{base}}_{T+1:T+H};
            Z^{\mathrm{ts}}_{T+1:T+H}
          \right]
        \right)
      \right)
    \in \mathbb{R}^{H\times D}
    \label{eq:future_covariate_residual}
\end{equation}
The final future states are:
\begin{equation}
    \widehat{Z}_{T+1:T+H}
    = \widetilde{Z}^{\mathrm{base}}_{T+1:T+H}
      + \Delta^{\mathrm{future}}_{T+1:T+H}
    \label{eq:final_state_prediction}
\end{equation}
Without future covariates, the base prediction is used directly. The resulting
future-state predictions receive latent-space supervision in Stage~1 and are
decoded into observations in Stage~2.

\subsection{Two-Stage Learning Objectives}
\label{sec:two_stage_learning}

\textbf{Stage~1: Latent Dynamics Learning.}
The predictor outputs future-state predictions $\widehat{Z}$, supervised by
states $Z$ obtained by encoding the ground-truth future observations with the shared
TS encoder. Inspired by LatentTSF~\citep{yang2026latentsf}, we match these states
using an element-wise $L_1$ loss and cosine similarity:
\begin{equation}
    \mathcal{L}_{\mathrm{latent}}
    =
    \left\|
        \widehat{Z}-\operatorname{sg}(Z)
    \right\|_1,
    \qquad
    \mathcal{L}_{\mathrm{sim}}
    =
    1-\cos\!\left(
        \widehat{Z},
        \operatorname{sg}(Z)
    \right)
    \label{eq:state_matching}
\end{equation}
Here, $\operatorname{sg}(\cdot)$ stops gradients; cosine similarity is
computed over flattened future state sequences.

To prevent representation collapse during state matching, we adopt the variance
and covariance regularizers from VICReg~\citep{bardes2022vicreg} for both
predicted and encoded future states, with statistics computed across samples
and forecast steps:
\begin{equation}
\begin{gathered}
\mathcal{V}(A)
=
\frac{1}{D}\sum_{d=1}^{D}
\left[
1-\sqrt{\operatorname{Var}(A_{:,d})+\epsilon}
\right]_{+},
\qquad
\mathcal{R}_{\mathrm{cov}}(A)
=
\frac{1}{D}\sum_{i\neq j}
[\operatorname{Cov}(A)]_{ij}^{2},
\\
\mathcal{L}_{\mathrm{VICReg}}
=
\frac{1}{2}
\sum_{A\in\{Z,\widehat{Z}\}}
\left[
\mathcal{V}(A)
+
\lambda_{\mathrm{cov}}
\mathcal{R}_{\mathrm{cov}}(A)
\right]
\end{gathered}
\label{eq:vicreg}
\end{equation}
We jointly optimize the learnable encoding modules and predictor with the
following objective:
\begin{equation}
    \mathcal{L}_{\mathrm{Stage1}}
    =
    \lambda_{\mathrm{sim}}\mathcal{L}_{\mathrm{sim}}
    +
    \lambda_{\mathrm{latent}}\mathcal{L}_{\mathrm{latent}}
    +
    \lambda_{\mathrm{VIC}}\mathcal{L}_{\mathrm{VICReg}}
    \label{eq:stage1_objective}
\end{equation}

\textbf{Stage~2: Observation Decoding.}
With the encoders and predictor frozen, $\widehat{Z}$ and $Z$ are passed through
the same TS decoder. Both outputs are supervised by the ground-truth future
observations $x$, yielding prediction and reconstruction losses that update
only the decoder:
\begin{equation}
    \mathcal{L}_{\mathrm{Stage2}}
    =
    \underbrace{
        \left\|
            x-D_{\phi}(\widehat{Z})
        \right\|_1
    }_{\mathcal{L}_{\mathrm{pred}}}
    +
    \alpha
    \underbrace{
        \left\|
            x-D_{\phi}(Z)
        \right\|_1
    }_{\mathcal{L}_{\mathrm{recon}}}
    \label{eq:stage2_objective}
\end{equation}
The prediction term adapts the decoder to states available at inference;
the reconstruction term provides additional supervision from states encoded
from the ground-truth future observations.

\section{Experiments}
\label{sec:experiments}

\subsection{Experimental Settings}
\label{sec:experimental_settings}

\noindent\textbf{Datasets.}
We evaluate WorldTS on 21 real-world datasets: 12 numerical-covariate
datasets used by DAG~\citep{qiu2026dag}, eight text-covariate Time-MMD
datasets~\citep{liu2024timemmd}, and the image-covariate MMSP solar-power
dataset~\citep{ma2024fusionsf}. See the appendix for details.

\noindent\textbf{Baselines.}
For numerical-covariate forecasting, we compare WorldTS with ten baselines, covering dedicated covariate-aware methods and general forecasting backbones,
including KITE~\citep{cheng2026kite} and DUET~\citep{qiu2025duet}.
For text-covariate forecasting, we compare with nine baselines covering
time-series-only and multimodal methods, including
PatchTST~\citep{nie2023patchtst} and VoT~\citep{wang2026vot}.
For image-covariate forecasting, following UniCA~\citep{han2026unica},
we compare with thirteen baselines spanning specialized forecasters,
pretrained time-series models, multimodal methods, and UniCA-adapted models,
including TTM-R2~\citep{ekambaram2024ttm} and FusionSF~\citep{ma2024fusionsf}.
The full baseline list and descriptions are provided in the appendix.



\noindent\textbf{Implementation Details.}
We implement WorldTS in PyTorch on NVIDIA A800 80GB GPUs and report MSE
and MAE. For numerical-covariate, text-covariate, and image-covariate \mbox{forecasting},
we follow the experimental settings of DAG~\citep{qiu2026dag},
VoT~\citep{wang2026vot}, and UniCA~\citep{han2026unica}, respectively.
Dataset-specific lookback windows and prediction \mbox{horizons} are detailed
in Appendix~\ref{app:implementation_details}.
``Drop Last'' is disabled at test time for all three settings to retain
the final incomplete batch~\citep{qiu2024tfb}.

\subsection{Main Findings}
\label{sec:main_results}

\renewcommand{\arraystretch}{1.15}
\begin{table*}[!t]
\centering
\caption{Average results over short and long forecasting horizons on the 12 numerical-covariate datasets. Baseline results are from DAG~\citep{qiu2026dag} and KITE~\citep{cheng2026kite} under the same evaluation protocol. Full results: Table~\ref{tab:numerical_full} (Appendix~\ref{app:full_numerical_results}). \bestresult{Red}: best; \secondresult{blue}: second best.}
\label{tab:numerical_main}
\setlength{\tabcolsep}{4pt}
\resizebox{\textwidth}{!}{%
\begin{tabular}{c|*{10}{cc|}cc}
\toprule
\multicolumn{1}{c|}{\raisebox{1.1ex}{Models}}
& \multicolumn{2}{c}{\shortstack{\textbf{WorldTS}\\(ours)}}
& \multicolumn{2}{c}{\shortstack{KITE\\(2026)}}
& \multicolumn{2}{c}{\shortstack{DAG\\(2026)}}
& \multicolumn{2}{c}{\shortstack{GCGNet\\(2026)}}
& \multicolumn{2}{c}{\shortstack{DUET\\(2025)}}
& \multicolumn{2}{c}{\shortstack{CrossLinear\\(2025)}}
& \multicolumn{2}{c}{\shortstack{Amplifier\\(2025)}}
& \multicolumn{2}{c}{\shortstack{TimeKAN\\(2025)}}
& \multicolumn{2}{c}{\shortstack{TimeXer\\(2024)}}
& \multicolumn{2}{c}{\shortstack{TiDE\\(2023)}}
& \multicolumn{2}{c}{\shortstack{TFT\\(2021)}} \\
\multicolumn{1}{c|}{Metrics}
& \multicolumn{1}{c}{mse} & \multicolumn{1}{c}{mae}
& \multicolumn{1}{c}{mse} & \multicolumn{1}{c}{mae}
& \multicolumn{1}{c}{mse} & \multicolumn{1}{c}{mae}
& \multicolumn{1}{c}{mse} & \multicolumn{1}{c}{mae}
& \multicolumn{1}{c}{mse} & \multicolumn{1}{c}{mae}
& \multicolumn{1}{c}{mse} & \multicolumn{1}{c}{mae}
& \multicolumn{1}{c}{mse} & \multicolumn{1}{c}{mae}
& \multicolumn{1}{c}{mse} & \multicolumn{1}{c}{mae}
& \multicolumn{1}{c}{mse} & \multicolumn{1}{c}{mae}
& \multicolumn{1}{c}{mse} & \multicolumn{1}{c}{mae}
& \multicolumn{1}{c}{mse} & \multicolumn{1}{c}{mae} \\
\midrule
NP
& \bestresult{0.297} & \bestresult{0.319} & \secondresult{0.325} & \secondresult{0.323}
& 0.362 & 0.344 & 0.370 & 0.348 & 0.411 & 0.407 & 0.371 & 0.387
& 0.420 & 0.418 & 0.405 & 0.419 & 0.418 & 0.371 & 0.443 & 0.400
& 0.379 & 0.375 \\
\midrule
PJM
& \bestresult{0.087} & \bestresult{0.178} & 0.096 & \bestresult{0.178}
& \secondresult{0.093} & \secondresult{0.180} & 0.095 & 0.187 & 0.102 & 0.197 & 0.112 & 0.223
& 0.137 & 0.246 & 0.139 & 0.262 & 0.108 & 0.198 & 0.142 & 0.246
& 0.114 & 0.207 \\
\midrule
BE
& \bestresult{0.393} & \bestresult{0.260} & 0.427 & 0.286
& \secondresult{0.423} & \secondresult{0.279} & 0.431 & 0.294 & 0.514 & 0.354 & 0.479 & 0.337
& 0.559 & 0.413 & 0.548 & 0.407 & 0.452 & 0.290 & 0.498 & 0.325
& 0.454 & 0.291 \\
\midrule
FR
& \bestresult{0.384} & \bestresult{0.210} & \secondresult{0.387} & 0.225
& 0.414 & \secondresult{0.219} & 0.415 & 0.234 & 0.495 & 0.327 & 0.482 & 0.298
& 0.554 & 0.408 & 0.547 & 0.374 & 0.427 & 0.241 & 0.484 & 0.281
& 0.504 & 0.257 \\
\midrule
DE
& \bestresult{0.340} & \bestresult{0.353} & \secondresult{0.350} & \secondresult{0.370}
& 0.370 & \secondresult{0.370} & 0.401 & 0.389 & 0.482 & 0.430 & 0.485 & 0.452
& 0.473 & 0.441 & 0.473 & 0.445 & 0.475 & 0.418 & 0.499 & 0.447
& 0.489 & 0.446 \\
\midrule
Energy
& \bestresult{0.091} & \bestresult{0.226} & \secondresult{0.111} & \secondresult{0.257}
& 0.124 & 0.267 & 0.131 & 0.277 & 0.203 & 0.367 & 0.239 & 0.401
& 0.233 & 0.389 & 0.218 & 0.381 & 0.163 & 0.315 & 0.153 & 0.302
& 0.130 & 0.283 \\
\midrule
Sdwpfm1
& \bestresult{0.391} & \bestresult{0.412} & \secondresult{0.417} & \secondresult{0.451}
& 0.423 & 0.461 & 0.424 & 0.457 & 0.599 & 0.570 & 0.426 & 0.502
& 0.437 & 0.490 & 0.447 & 0.534 & 0.701 & 0.609 & 0.483 & 0.507
& 0.482 & 0.474 \\
\midrule
Sdwpfm2
& \bestresult{0.446} & \bestresult{0.450} & \secondresult{0.475} & \secondresult{0.483}
& 0.477 & 0.485 & \secondresult{0.475} & 0.486 & 0.514 & 0.490 & 0.533 & 0.573
& 0.491 & 0.512 & 0.497 & 0.564 & 0.803 & 0.653 & 0.486 & 0.516
& 0.476 & 0.488 \\
\midrule
Sdwpfh1
& \bestresult{0.376} & \bestresult{0.431} & \secondresult{0.441} & 0.491
& 0.448 & \secondresult{0.486} & 0.450 & 0.500 & 0.539 & 0.516 & 0.557 & 0.593
& 0.537 & 0.598 & 0.577 & 0.638 & 0.746 & 0.643 & 0.453 & 0.508
& 0.479 & 0.491 \\
\midrule
Sdwpfh2
& \bestresult{0.419} & \bestresult{0.461} & \secondresult{0.500} & 0.526
& 0.523 & 0.530 & 0.520 & 0.536 & 0.647 & 0.566 & 0.538 & 0.574
& 0.521 & 0.581 & 0.647 & 0.672 & 0.891 & 0.719 & 0.599 & 0.583
& 0.566 & \secondresult{0.521} \\
\midrule
Colbun
& \bestresult{0.088} & \bestresult{0.146} & \bestresult{0.088} & 0.172
& \secondresult{0.098} & \secondresult{0.154} & 0.107 & 0.175 & 0.198 & 0.266 & 0.126 & 0.195
& 0.173 & 0.246 & 0.128 & 0.175 & 0.145 & 0.235 & 0.164 & 0.227
& 0.238 & 0.297 \\
\midrule
Rapel
& \bestresult{0.228} & \bestresult{0.259} & 0.244 & \secondresult{0.294}
& \secondresult{0.230} & 0.305 & 0.306 & 0.307 & 0.269 & 0.326 & 0.252 & 0.313
& 0.257 & 0.321 & 0.249 & 0.311 & 0.344 & 0.362 & 0.320 & 0.351
& 0.305 & 0.333 \\
\midrule
1st Count
& \bestresult{12} & \bestresult{12} & 1 & 1 & 0 & 0 & 0 & 0 & 0 & 0
& 0 & 0 & 0 & 0 & 0 & 0 & 0 & 0 & 0 & 0 & 0 & 0 \\
\bottomrule
\end{tabular}%
}
\end{table*}
\noindent\textbf{Numerical-Covariate Forecasting.}
As shown in Table~\ref{tab:numerical_main}, WorldTS achieves the lowest MSE
and MAE across all 12 datasets. On Energy, it reduces MSE by 18.0\% relative
to the strongest baseline. These results show its effectiveness
across diverse numerical-covariate forecasting tasks.

\renewcommand{\arraystretch}{1.1}
\begin{table*}[!t]
\centering
\caption{Forecasting results on eight Time-MMD datasets with a prediction horizon of 12. Baseline results are taken from VoT~\citep{wang2026vot} under the same evaluation setting. }
\label{tab:timemmd_main}
\resizebox{\textwidth}{!}{%
\begin{tabular}{c|cc|*{6}{cc}|*{3}{cc}}
\toprule
\multicolumn{1}{c|}{\raisebox{1.1ex}{Models}}
& \multicolumn{2}{c}{\shortstack{\textbf{WorldTS}\\(ours)}}
& \multicolumn{2}{c}{\shortstack{VoT\\(2026)}}
& \multicolumn{2}{c}{\shortstack{TaTS\\(2026)}}
& \multicolumn{2}{c}{\shortstack{GPT4TS\\(2025)}}
& \multicolumn{2}{c}{\shortstack{Time-VLM\\(2025)}}
& \multicolumn{2}{c}{\shortstack{CALF\\(2025)}}
& \multicolumn{2}{c}{\shortstack{GPT4MTS\\(2024)}}
& \multicolumn{2}{c}{\shortstack{RaFT\\(2025)}}
& \multicolumn{2}{c}{\shortstack{iTransformer\\(2024)}}
& \multicolumn{2}{c}{\shortstack{PatchTST\\(2023)}} \\
\multicolumn{1}{c|}{Metrics}
& \multicolumn{1}{c}{mse} & \multicolumn{1}{c}{mae}
& \multicolumn{1}{c}{mse} & \multicolumn{1}{c}{mae}
& \multicolumn{1}{c}{mse} & \multicolumn{1}{c}{mae}
& \multicolumn{1}{c}{mse} & \multicolumn{1}{c}{mae}
& \multicolumn{1}{c}{mse} & \multicolumn{1}{c}{mae}
& \multicolumn{1}{c}{mse} & \multicolumn{1}{c}{mae}
& \multicolumn{1}{c}{mse} & \multicolumn{1}{c}{mae}
& \multicolumn{1}{c}{mse} & \multicolumn{1}{c}{mae}
& \multicolumn{1}{c}{mse} & \multicolumn{1}{c}{mae}
& \multicolumn{1}{c}{mse} & \multicolumn{1}{c}{mae} \\
\midrule
Agriculture
& \bestresult{0.285} & \bestresult{0.338} & \secondresult{0.288} & 0.355
& 0.290 & 0.350 & 0.291 & \bestresult{0.338} & 0.322 & 0.359
& 0.314 & 0.355 & 0.301 & \secondresult{0.342} & 0.301 & 0.383 & 0.300 & 0.352 & 0.309 & 0.351 \\
\midrule
Climate
& \bestresult{1.072} & \bestresult{0.838} & \secondresult{1.088} & \secondresult{0.845}
& 1.179 & 0.885 & 1.171 & 0.883 & 1.203 & 0.896
& 1.177 & 0.883 & 1.152 & 0.876 & 1.295 & 0.930 & 1.131 & 0.864 & 1.168 & 0.877 \\
\midrule
Economy
& \bestresult{0.214} & \bestresult{0.363} & 0.223 & \secondresult{0.374}
& 0.229 & 0.381 & 0.233 & 0.386 & 0.245 & 0.396
& \secondresult{0.216} & \bestresult{0.363} & 0.226 & 0.379 & 0.280 & 0.425 & 0.230 & 0.385 & 0.229 & 0.382 \\
\midrule
Energy
& \bestresult{0.090} & \bestresult{0.201} & \secondresult{0.091} & \secondresult{0.218}
& 0.105 & 0.232 & 0.111 & 0.243 & 0.114 & 0.253
& 0.102 & 0.224 & 0.111 & 0.244 & 0.123 & 0.255 & 0.121 & 0.258 & 0.107 & 0.235 \\
\midrule
Health
& \bestresult{0.778} & \bestresult{0.542} & 0.898 & \secondresult{0.596}
& 0.899 & 0.612 & \secondresult{0.854} & 0.629 & 1.198 & 0.727
& 0.964 & 0.609 & 0.985 & 0.658 & 1.297 & 0.769 & 1.261 & 0.746 & 1.006 & 0.650 \\
\midrule
Security
& \bestresult{74.325} & \bestresult{3.979} & 75.931 & 4.153
& 77.656 & 4.239 & 76.965 & 4.219 & 80.767 & 4.438
& \secondresult{74.631} & \secondresult{4.113} & 78.041 & 4.316 & 83.911 & 4.712 & 82.644 & 4.493 & 77.815 & 4.254 \\
\midrule
SocialGood
& \secondresult{0.962} & \bestresult{0.421} & \bestresult{0.925} & \secondresult{0.438}
& 1.053 & 0.474 & 1.167 & 0.608 & 1.005 & 0.505
& 0.991 & 0.439 & 1.093 & 0.470 & 1.151 & 0.553 & 1.153 & 0.574 & 1.263 & 0.643 \\
\midrule
Traffic
& \secondresult{0.185} & \bestresult{0.238} & \bestresult{0.181} & \secondresult{0.239}
& 0.189 & 0.242 & 0.211 & 0.260 & 0.222 & 0.322
& 0.193 & 0.243 & 0.218 & 0.268 & 0.300 & 0.387 & 0.205 & 0.250 & 0.188 & \secondresult{0.239} \\
\midrule
1st Count
& \bestresult{6} & \bestresult{8} & 2 & 0 & 0 & 0 & 0 & 1 & 0 & 0
& 0 & 1 & 0 & 0 & 0 & 0 & 0 & 0 & 0 & 0 \\
\bottomrule
\end{tabular}%
}
\end{table*}

\noindent\textbf{Text-Covariate Forecasting.}
Across the eight Time-MMD datasets, WorldTS ranks first in 14 of 16
dataset--metric comparisons against the time-series-only and multimodal
baselines.
The largest gains are on Health: 8.9\% in MSE and 9.1\% in MAE
relative to the strongest baselines.

\begin{table*}[!t]
\centering
\renewcommand{\arraystretch}{1.1}
\caption{
Forecasting results on the MMSP with a prediction horizon of 24.
Baseline results are taken from UniCA~\citep{han2026unica} under the same evaluation setting.
}
\label{tab:mmsp_main}
\setlength{\tabcolsep}{4pt}
\resizebox{\textwidth}{!}{%
\begin{tabular}{c*{14}{c}}
\toprule
\multicolumn{1}{c}{Category}
& \multicolumn{1}{c}{Ours}
& \multicolumn{2}{c}{Multimodal}
& \multicolumn{5}{c}{Specialized Forecasters}
& \multicolumn{3}{c}{Pretrained Models}
& \multicolumn{3}{c}{UniCA-adapted} \\
\cmidrule(lr){2-2}
\cmidrule(lr){3-4}
\cmidrule(lr){5-9}
\cmidrule(lr){10-12}
\cmidrule(l){13-15}
Models
& \textbf{WorldTS}
& MM-TSF & FusionSF
& TFT & D.AR & NB.S & TiDE & P.TST
& TTM-R2 & Moirai & PFN-TS
& Chronos-Bolt & TimesFM & Time-MoE \\
\midrule
MSE
& \bestresult{0.061}
& 0.090 & 0.478
& \secondresult{0.067} & 0.097 & 0.107 & 0.297 & 0.662
& 0.095 & 0.206 & 0.633
& 0.090 & 0.098 & 0.653 \\
MAE
& \bestresult{0.171}
& 0.200 & 0.566
& \secondresult{0.177} & 0.216 & 0.219 & 0.438 & 0.682
& 0.263 & 0.378 & 0.711
& 0.193 & 0.229 & 0.765 \\
\bottomrule
\end{tabular}%
}
\end{table*}
\noindent\textbf{Image-Covariate Forecasting.}
Table~\ref{tab:mmsp_main} reports results on MMSP, where WorldTS achieves
the lowest MSE and MAE. Compared with TFT, the strongest baseline on both metrics, WorldTS reduces
MSE and MAE by approximately 9.0\% and 3.4\%, respectively. These results show the effectiveness of WorldTS in
leveraging numerical and visual covariates.

\subsection{Parameter Sensitivity}
\label{sec:parameter_sensitivity}

\begin{figure*}[t]
    \centering
    \includegraphics[width=\textwidth]
    {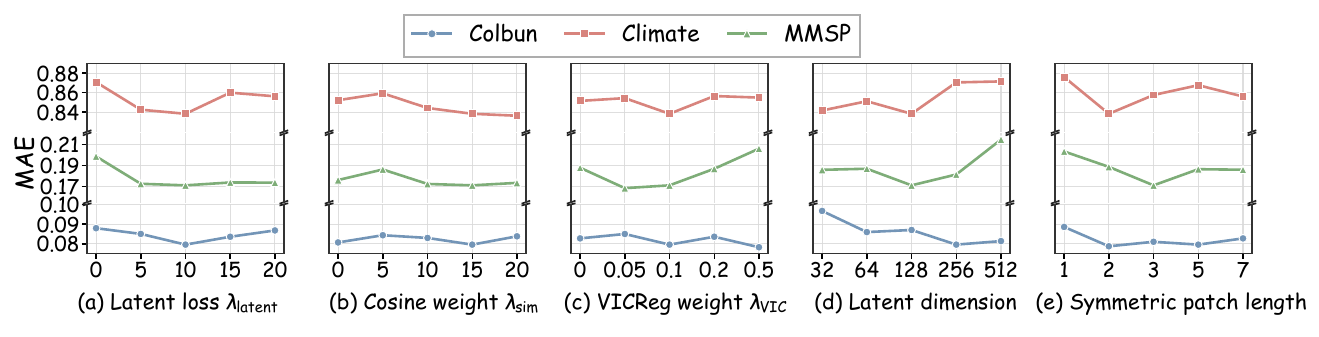}
    \caption{Parameter sensitivity of WorldTS. We vary the weights of three losses
($\lambda_{\mathrm{latent}}$, $\lambda_{\mathrm{sim}}$, and
$\lambda_{\mathrm{VIC}}$), the latent-state dimensionality, and the symmetric causal
patch length one at a time on Colbun, Climate, and MMSP, and report MAE.
Other settings remain fixed.}
    \label{fig:parameter_sensitivity}
\end{figure*}

We assess parameter sensitivity on Colbun, Climate, and MMSP by varying
one hyperparameter at a time. The results are shown in Figure~\ref{fig:parameter_sensitivity}. We make the following observations:
1) Setting any loss weight to zero increases MAE relative to its default
on all three datasets, indicating that each loss contributes to prediction
accuracy. 2) Among the tested values, $\lambda_{\text{latent}} = 10$ achieves the lowest MAE on all three datasets, while $\lambda_{\text{sim}} = 15$ and $\lambda_{\text{VIC}} = 0.1$ consistently achieve either the best or second-best performance.
Increasing $\lambda_{\mathrm{VIC}}$ to 0.5 increases the MMSP error markedly,
showing that stronger regularization does not necessarily improve performance.
3) Moderate latent-state dimensionalities perform better: 128 or 256 dimensions
yield the lowest MAE, with no consistent gains from larger dimensions.
4) Moderate causal patch lengths generally perform better, suggesting that
overly short or long temporal contexts can reduce forecasting accuracy.

\subsection{Ablation Study}
\label{sec:ablation}

\begin{table*}[t]
\centering
\caption{Ablation results on representative datasets with numerical, textual, and visual covariates.}
\label{tab:ablation_main}
\small
\setlength{\tabcolsep}{2.2pt}
\renewcommand{\arraystretch}{1.12}
\resizebox{\textwidth}{!}{%
\begin{tabular}{lccccccc}
\toprule
\shortstack{Dataset\\[-1pt]{\scriptsize (MSE / MAE)}}
& Full
& w/o $\mathcal{L}_{\mathrm{latent}}$
& w/o $\mathcal{L}_{\mathrm{sim}}$
& w/o $\mathcal{L}_{\mathrm{VICReg}}$
& Pointwise MLP
& One-stage MAE
& w/o Covariates \\
\midrule
Colbun
& \textbf{0.052 / 0.080}
& 0.056 / 0.088
& 0.055 / 0.081
& 0.053 / 0.083
& 0.066 / 0.091
& 0.059 / 0.096
& 0.073 / 0.088 \\
Climate
& \textbf{1.072 / 0.838}
& 1.145 / 0.871
& 1.089 / 0.852
& 1.111 / 0.852
& 1.086 / 0.844
& 1.149 / 0.875
& 1.134 / 0.869 \\
MMSP
& \textbf{0.061 / 0.171}
& 0.078 / 0.199
& 0.063 / 0.176
& 0.068 / 0.188
& 0.065 / 0.187
& 0.085 / 0.217
& 0.101 / 0.279 \\
\bottomrule
\end{tabular}%
}
\end{table*}
We compare the full model with six ablation variants and report the results in
Table~\ref{tab:ablation_main} and the protocols in Appendix~\ref{app:ablation_protocols}.
We make the following observations:
1) Removing any loss degrades forecasting performance on all three datasets,
indicating that the three losses complement each other and perform better
when used jointly.
2) Replacing causal state mappings with pointwise MLPs consistently increases
forecasting errors, suggesting that aggregating local temporal information
through overlapping causal patches helps learn effective predicted states.
3) One-stage training using only observation-space MAE underperforms relative to the full
model, suggesting that explicit latent-space supervision helps improve forecasting accuracy.
4) Removing covariates increases errors across datasets, especially on MMSP,
highlighting the benefits of using external information.

















\section{Conclusion}

Existing covariate-aware forecasting methods mainly use external information
for observation-space forecasting, while the role of multimodal covariates in
future-state prediction remains underexplored. Inspired by latent-space prediction
in world models, WorldTS uses direct future-state supervision and two-stage
training to learn latent space dynamics conditioned on multimodal covariates and to decode
predicted future states into future observations. Experiments on 21 real-world
datasets offer insight into the properties and the effectiveness of WorldTS; further analyses suggest that
using covariates improves both forecasting accuracy and the quality of predicted future states.




\bibliography{references}
\bibliographystyle{worldts-preprint}

\newpage
\appendix
\raggedbottom
\section{Experimental Details}
\label{app:experimental_details}

\subsection{Datasets}
\label{app:datasets}

\paragraph{Numerical-covariate datasets.}
For numerical-covariate forecasting, we use the 12 TSF-X datasets adopted by
DAG~\citep{qiu2026dag}. Five of them are electricity-market benchmarks. NP
uses the Nord Pool electricity price as the forecasting target and provides
grid-load and wind-power forecasts as external variables. PJM focuses on the
zonal price in the Commonwealth Edison area, accompanied by system-level and
regional load forecasts. BE, FR, and DE contain electricity prices from the
Belgian, French, and German markets, respectively; their covariates comprise
market-specific load and generation forecasts, including wind and solar
generation where available. The Energy dataset describes the hourly
generation mix of the Chilean power system. We forecast thermoelectric output
using generation from the other energy sources as covariates. Colbun and Rapel
are daily hydropower datasets from two Chilean reservoirs, where the future
water level is predicted from precipitation and tributary-inflow measurements.
Finally, Sdwpfm1, Sdwpfm2, Sdwpfh1, and Sdwpfh2 are constructed from two wind
turbines at two temporal resolutions. Their target is active power, and the
external variables are ERA5 meteorological measurements, including
temperature, surface pressure, relative humidity, wind conditions, and
precipitation.

\paragraph{Text-covariate datasets.}
For text-covariate forecasting, we use eight domains from
Time-MMD~\citep{liu2024timemmd}: Agriculture, Climate, Economy, Energy,
Health, Security, Social Good, and Traffic. Each domain associates a numerical target
series with time-relevant reports or other textual records. Agriculture pairs
broiler-market prices with USDA market reports; Climate combines NOAA drought indices with monthly climate reports; and
Economy links international trade measurements with releases on trade and
economic indicators. The Energy domain augments gasoline-price observations
with reports published by the U.S. Energy Information Administration, while
Health combines influenza-like-illness statistics with influenza surveillance
material. Security concerns disaster and emergency records and is supplemented
with disaster-related reports. Social Good uses U.S. unemployment statistics
together with employment and labor-force publications, and Traffic pairs
traffic-volume measurements with corresponding transportation reports. These
datasets evaluate whether temporally associated language can provide useful
context for predicting the evolution of a numerical target.

\paragraph{Image-covariate dataset.}
For image-covariate forecasting, we use the Multimodal Solar Power (MMSP)
dataset~\citep{ma2024fusionsf}, which contains approximately one and a half
years of hourly observations from 88 photovoltaic plants. Following
FusionSF~\citep{ma2024fusionsf} and UniCA~\citep{han2026unica}, we evaluate
the first 10 plants. The target is future solar-power output, and the auxiliary
information includes aligned satellite imagery, numerical weather predictions,
and plant coordinates. The satellite observations provide spatial weather
information beyond the historical power sequence, making MMSP a natural
benchmark for visual and numerical covariates.

\subsection{Baselines}
\label{app:baselines}

To ensure a comprehensive and representative comparison, we compare WorldTS against a broad set of competitive baselines across numerical, text, and image covariates. These baselines include dedicated covariate-aware models, strong general-purpose forecasters, multimodal forecasting methods, and pretrained or foundation models, covering the major forecasting paradigms.

\paragraph{Numerical-covariate baselines.}
We compare WorldTS with methods covering several approaches to numerical
covariate modeling. KITE~\citep{cheng2026kite} uses external knowledge to guide probabilistic
forecasting, whereas DAG~\citep{qiu2026dag} explicitly models the relations between endogenous
and exogenous series and between historical and forecast-period
variables. GCGNet~\citep{li2026gcgnet} represents cross-series
dependencies through a graph-based generative model, and
TimeXer~\citep{wang2024timexer} introduces variate-level representations through which a Transformer
can incorporate exogenous series. TFT~\citep{lim2021tft} combines variable
selection, gating, and temporal attention to process observed and known-future
inputs, while TiDE~\citep{das2023tide} uses a dense encoder--decoder architecture with explicit
access to dynamic covariates. To provide broader comparisons with general-purpose forecasting approaches, we additionally include DUET~\citep{qiu2025duet}, which learns relationships among dynamically clustered channels; CrossLinear~\citep{zhou2025crosslinear}, which constructs cross-correlation representations across variables; Amplifier~\citep{fei2025amplifier}, which emphasizes informative low-energy signal components; and TimeKAN~\citep{huang2025timekan}, which models temporal patterns after frequency decomposition. Together, these baselines cover both dedicated exogenous-variable models and competitive general-purpose time-series forecasting architectures.

\paragraph{Text-covariate baselines.}
We first consider three methods that forecast from numerical histories without
using the paired reports. RaFT~\citep{han2025raft} augments prediction by retrieving relevant
training sequences, iTransformer~\citep{liu2024itransformer} treats variables as tokens, and PatchTST~\citep{nie2023patchtst}
models the input through channel-independent temporal
patches. We also compare with a range of multimodal time-series forecasting methods. VoT~\citep{wang2026vot} extracts event-level evidence from text and
aligns its textual and numerical predictions at multiple levels, whereas TaTS~\citep{li2026tats}
organizes time-series-paired documents into a temporally coherent textual
representation. GPT4TS~\citep{chang2025llm4ts} adapts a pretrained
language model to numerical time-series patterns, and GPT4MTS~\citep{jia2024gpt4mts} formulates
multimodal forecasting through prompt-based interaction with a large language
model. Time-VLM~\citep{zhong2025timevlm} uses a vision--language
model to augment temporal prediction, while CALF~\citep{liu2025calf} learns cross-modal alignment
through targeted fine-tuning of a pretrained language
model. 

\paragraph{Image-covariate baselines.}
Following the MMSP comparison protocol of UniCA~\citep{han2026unica}, we include
four families of methods. The specialized forecasters comprise PatchTST~\citep{nie2023patchtst},
N-BEATSx~\citep{nbeatsx}, DeepAR~\citep{salinas2020deepar}, TFT~\citep{lim2021tft}, and TiDE~\citep{das2023tide}. They span target-only patch modeling,
autoregressive probabilistic prediction, and architectures that can consume
structured historical or future covariates. TTM-R2~\citep{ekambaram2024ttm}, Moirai~\citep{Woo2024Moirai},
and TabPFN-TS~\citep{hoo2024tabular} represent pretrained forecasting models with different
MLP-mixing, Transformer, and tabular pretraining strategies. FusionSF~\citep{ma2024fusionsf} and MM-TSF~\citep{liu2024timemmd} are
included as explicit multimodal forecasters. Finally, we report Chronos-Bolt~\citep{ansari2024chronos}, TimesFM~\citep{das2024timesfm}, and
Time-MoE~\citep{shi2025timemoe} equipped with UniCA, which supplies a common covariate-adaptation
interface to otherwise heterogeneous time-series foundation
models.
This collection contrasts WorldTS with task-specific models, pretrained
forecasters, multimodal forecasting methods, and covariate-adapted foundation
models under the same evaluation setting.

\subsection{Implementation Details}
\label{app:implementation_details}

\begin{figure}[t]
    \centering
    \includegraphics[width=0.7\linewidth]{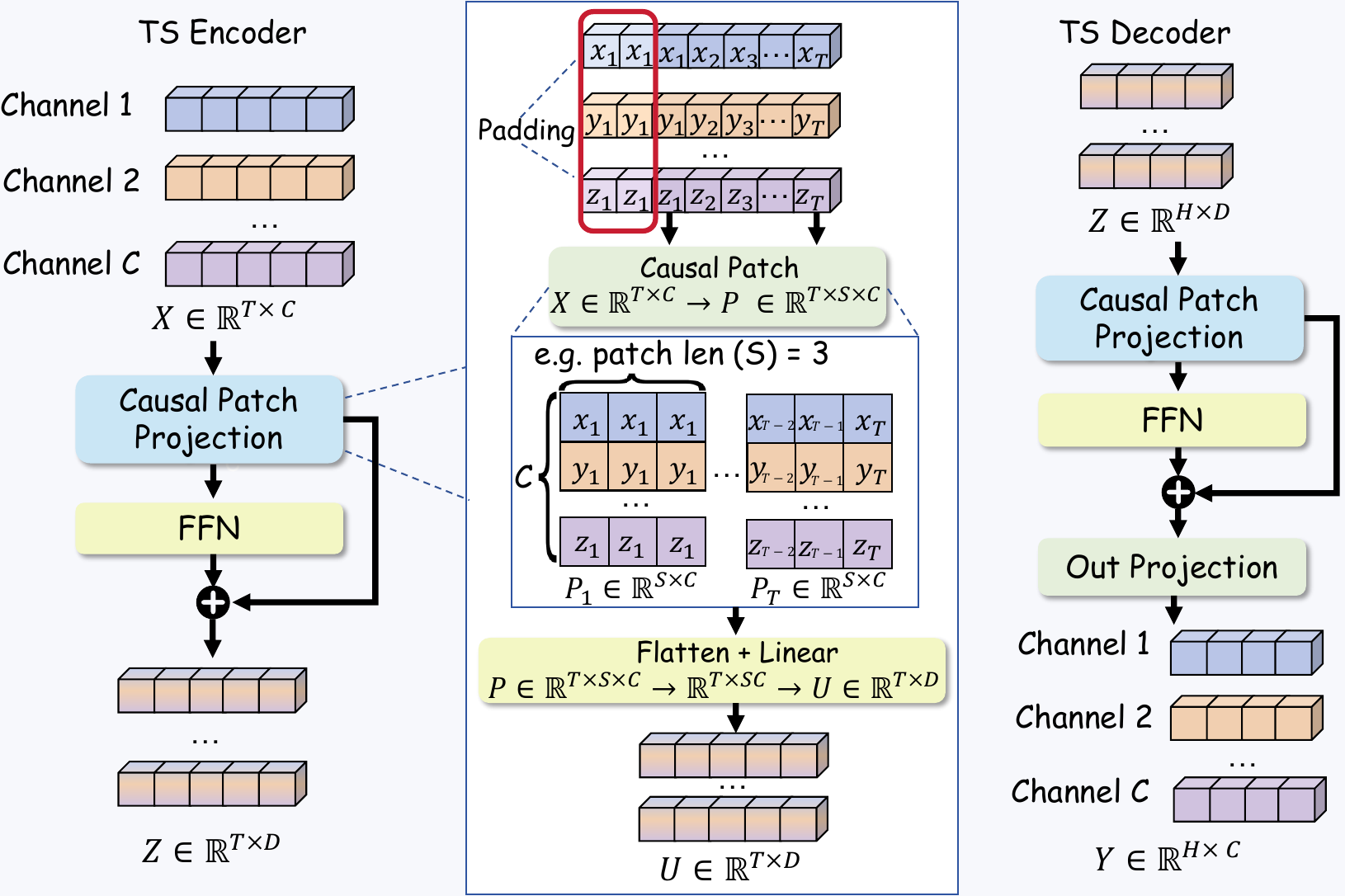}
    \caption{TS Encoder and Decoder. Overlapping causal observation patches
    yield time-aligned states, while local state patches reconstruct target
    observations.}
    \label{fig:state_encoder_decoder}
\end{figure}

\paragraph{Data splits and environment.} Following DAG~\citep{qiu2026dag}, the numerical-covariate datasets are chronologically split into 70\% training, 10\% validation, and 20\% test observations. Following VoT~\citep{wang2026vot}, Time-MMD adopts the same 70\%/10\%/20\% chronological training/validation/test split. Following UniCA~\citep{han2026unica}, MMSP is chronologically split into 80\% training, 10\% validation, and 10\% test observations. All experiments are implemented in PyTorch and run on NVIDIA A800 80GB GPUs. The numerical-covariate experiments use Python 3.8, while the text- and image-covariate experiments use Python 3.10.
Both stages use Adam, with seed 2021 for numerical-covariate experiments
and 2025 for Time-MMD and MMSP.

\begin{algorithm}[t]
\caption{Pointwise Encoder--Decoder Ablation}
\label{alg:pointwise}
\begin{algorithmic}[1]
\STATE \textbf{Input:} Training set $\mathcal{D}$; historical and future observations
$x_{1:T}$ and $x_{T+1:T+H}$; multimodal covariates $\mathcal{C}$;
pointwise MLP encoders $\bar{E}_{\theta}$ and
$\bar{E}^{\mathrm{ts}}_{\theta_{\mathrm{ts}}}$;
unchanged auxiliary encoders
$\{E_{\theta_m}^{m}\}_{m\in\{\mathrm{txt},\mathrm{img}\}}$;
state predictor $P_{\psi}$; pointwise MLP decoder $\bar{D}_{\phi}$;
learning rates $\eta_1,\eta_2$; epochs $N_1,N_2$

\STATE \textbf{Output:} Forecast $\widehat{x}_{T+1:T+H}$

\STATE \textbf{Stage 1: Latent Dynamics Learning}
\FOR{$i=1$ to $N_1$}
    \FOR{$(x_{1:T},x_{T+1:T+H},\mathcal{C})\in\mathcal{D}$}
        \STATE $Z_{1:T}\leftarrow \bar{E}_{\theta}(x_{1:T})$
        \STATE $Z_{T+1:T+H}\leftarrow
        \bar{E}_{\theta}(x_{T+1:T+H})$
        \STATE Encode $\{Z^m\}_{m\in\mathcal{M}}$ from $\mathcal{C}$,
        using $\bar{E}^{\mathrm{ts}}_{\theta_{\mathrm{ts}}}$ for numerical covariates
        \STATE $\widehat{Z}_{T+1:T+H}
        \leftarrow
        P_{\psi}\!\left(
        Z_{1:T},
        \{Z^m\}_{m\in\mathcal{M}}
        \right)$
        \STATE $\mathcal{L}\leftarrow\mathcal{L}_{\mathrm{Stage1}}$
        \STATE Update the encoders and $P_{\psi}$ using
        $\eta_1\nabla\mathcal{L}$
    \ENDFOR
\ENDFOR

\STATE Freeze the encoders and $P_{\psi}$

\STATE \textbf{Stage 2: Observation Decoding}
\FOR{$i=1$ to $N_2$}
    \FOR{$(x_{1:T},x_{T+1:T+H},\mathcal{C})\in\mathcal{D}$}
        \STATE Compute $\widehat{Z}_{T+1:T+H}$ and
        $Z_{T+1:T+H}$ with the frozen modules
        \STATE $\widehat{x}_{T+1:T+H}
        \leftarrow
        \bar{D}_{\phi}(\widehat{Z}_{T+1:T+H})$
        \STATE $\widetilde{x}_{T+1:T+H}
        \leftarrow
        \bar{D}_{\phi}(Z_{T+1:T+H})$
        \STATE $\mathcal{L}\leftarrow\mathcal{L}_{\mathrm{Stage2}}$
        \STATE Update $\bar{D}_{\phi}$ using
        $\eta_2\nabla\mathcal{L}$
    \ENDFOR
\ENDFOR

\STATE \textbf{Inference:} compute
$\widehat{Z}_{T+1:T+H}$ from $x_{1:T}$ and $\mathcal{C}$
\STATE \textbf{return}
$\widehat{x}_{T+1:T+H}
=
\bar{D}_{\phi}(\widehat{Z}_{T+1:T+H})$
\end{algorithmic}
\end{algorithm}

\paragraph{Implementation details.}
Following DAG~\citep{qiu2026dag}, we evaluate numerical-covariate forecasting
under both short-term and long-term settings. For Colbun and Rapel,
short-term forecasting uses a lookback window of 60 steps with a prediction
horizon of 10 steps, while long-term forecasting uses a lookback window of
180 steps with a prediction horizon of 30 steps. For all other datasets, short-term forecasting uses a lookback window of 168 steps with a prediction horizon of 24 steps, while long-term forecasting uses a lookback window of 720 steps with a prediction horizon of 360 steps. Following VoT~\citep{wang2026vot}, we use a lookback window of 36 steps for
Energy and Health and 8 steps for the remaining six Time-MMD datasets.
We select a prediction horizon of 12 steps for all eight datasets. Following UniCA~\citep{han2026unica}, we evaluate image-covariate forecasting on MMSP using a 2,048-step lookback window and a 24-step prediction horizon.

\paragraph{Modality encoders.}
The text and image encoders in Eq.~\ref{eq:historical_target_state} are
instantiated as: 
\begin{equation}
    \begin{aligned}
        Z^{\mathrm{txt}}_{1:T}
        &= E_{\theta_{\mathrm{txt}}}^{\mathrm{txt}}
           \!\left(c^{\mathrm{txt}}_{1:T}\right)
         = \operatorname{MLP}_{\mathrm{txt}}
           \!\left(
             \operatorname{MeanPool}\!\left(
               \operatorname{Embed}_{\mathrm{GPT2}}
               \!\left(c^{\mathrm{txt}}_{1:T}\right)
             \right)
           \right), \\
        Z^{\mathrm{img}}_{1:T}
        &= E_{\theta_{\mathrm{img}}}^{\mathrm{img}}
           \!\left(c^{\mathrm{img}}_{1:T}\right)
         = \operatorname{MLP}_{\mathrm{img}}
           \!\left(
             \operatorname{Linear}_{\mathrm{img}}\!\left(
               \operatorname{CNN}\!\left(c^{\mathrm{img}}_{1:T}\right)
             \right)
           \right)
    \end{aligned}
    \label{eq:modality_encoder_implementation}
\end{equation}
Following VoT~\citep{wang2026vot}, the text branch uses the input embedding
layer of a frozen GPT-2 model to obtain token-level representations and mean
pools the valid tokens at each time step. The pooled features are mapped by
$\operatorname{MLP}_{\mathrm{txt}}$ to the time-aligned text representations
$Z^{\mathrm{txt}}_{1:T}$. Following UniCA~\citep{han2026unica}, the image
branch uses a lightweight CNN to extract frame-wise spatial features.
$\operatorname{Linear}_{\mathrm{img}}$ converts the heterogeneous visual
features into compact continuous covariate representations, which are mapped
by $\operatorname{MLP}_{\mathrm{img}}$ to the time-aligned image
representations $Z^{\mathrm{img}}_{1:T}$.

\subsection{Ablation Protocols}
\label{app:ablation_protocols}

In the following algorithms, we abbreviate $c^m_{1:\tau_m}$ and
$Z^m_{1:\tau_m}$ as $c^m$ and $Z^m$, respectively.

\paragraph{Pointwise encoder--decoder.}
We replace the state encoder, numerical-covariate encoder, and state
decoder with two-layer pointwise MLPs. At each time step, the state encoder
maps an observation $x_t\in\mathbb{R}^{C}$ to a state
$Z_t\in\mathbb{R}^{D}$, while the decoder maps the
predicted state back to the observation space:
\begin{equation}
    Z_t
    = \bar{E}_{\theta}(x_t),
    \qquad
    \widehat{x}_t
    = \bar{D}_{\phi}\!\left(\widehat{Z}_t\right)
    \label{eq:pointwise_state_mappings}
\end{equation}
These mappings preserve the original input and output dimensions and share
parameters across time, but do not aggregate information from neighboring
time steps. The text and image encoders, state predictor, learning objectives,
and two-stage training protocol remain unchanged.
Algorithm~\ref{alg:pointwise} summarizes this variant.

\begin{algorithm}[t]
\caption{One-Stage Observation-Space MAE Training}
\label{alg:one_stage}
\begin{algorithmic}[1]
\STATE \textbf{Input:} Training set $\mathcal{D}$; historical and future observations
$x_{1:T}$ and $x_{T+1:T+H}$; multimodal covariates $\mathcal{C}$;
state encoder $E_{\theta}$;
modality encoders $\{E_{\theta_m}^{m}\}_{m\in\mathcal{M}}$;
state predictor $P_{\psi}$; state decoder $D_{\phi}$;
learning rate $\eta$; epochs $N_1+N_2$

\STATE \textbf{Output:} Forecast $\widehat{x}_{T+1:T+H}$

\STATE \textbf{End-to-end training:}
\FOR{$i=1$ to $N_1+N_2$}
    \FOR{$(x_{1:T},x_{T+1:T+H},\mathcal{C})\in\mathcal{D}$}
        \STATE $Z_{1:T}\leftarrow E_{\theta}(x_{1:T})$
        \STATE $\{Z^m\}_{m\in\mathcal{M}}
        \leftarrow
        \{E_{\theta_m}^{m}(c^m)\}_{m\in\mathcal{M}}$
        \STATE $\widehat{Z}_{T+1:T+H}
        \leftarrow
        P_{\psi}\!\left(
        Z_{1:T},
        \{Z^m\}_{m\in\mathcal{M}}
        \right)$
        \STATE $\widehat{x}_{T+1:T+H}
        \leftarrow
        D_{\phi}(\widehat{Z}_{T+1:T+H})$
        \STATE $\mathcal{L}_{\mathrm{obs}}
        \leftarrow
        \left\|
        \widehat{x}_{T+1:T+H}
        -
        x_{T+1:T+H}
        \right\|_1$
        \STATE Jointly update $\theta$, $\{\theta_m\}$,
        $\psi$, and $\phi$ using
        $\eta\nabla\mathcal{L}_{\mathrm{obs}}$
    \ENDFOR
\ENDFOR

\STATE \textbf{Inference:} compute
$\widehat{x}_{T+1:T+H}$ from $x_{1:T}$ and $\mathcal{C}$
\STATE \textbf{return} $\widehat{x}_{T+1:T+H}$
\end{algorithmic}
\end{algorithm}

\paragraph{One-stage observation-space training.}
To examine the effect of explicit predictive-state learning and the
two-stage strategy, we remove the future-state encoding branch and
its state-space objectives. The original causal patch encoder and decoder
are retained, and the complete forecasting path is optimized end to end
using only observation-space MAE. We train for $N_1+N_2$ epochs to match the
total number of training epochs of the full model. The resulting procedure is
given in Algorithm~\ref{alg:one_stage}.

For each loss ablation, only the corresponding coefficient
$\lambda_k$, $k\in\{\mathrm{latent},\mathrm{sim},\mathrm{VIC}\}$, is set to
zero. For the variant without covariates, we set
$\mathcal{C}=\varnothing$ while retaining the full architecture, objectives,
and two-stage training protocol.

\section{Full Numerical-Covariate Results}
\label{app:full_numerical_results}

Table~\ref{tab:numerical_full} reports the complete MSE and MAE results for all forecasting horizons on the 12 numerical-covariate datasets, together with the averages reported in the main text. WorldTS achieves the best or tied-best average performance on all datasets; for the few individual forecasting settings where it does not rank first, it typically remains second-best or within a competitive range.

\renewcommand{\arraystretch}{1.3}
\begin{table*}[t]
\centering
\caption{Full results on the 12 numerical-covariate datasets. \textit{Avg}
denotes the average result over the two forecasting horizons. Baseline results
follow the same evaluation protocol as Table~\ref{tab:numerical_main}. The best
results are in \bestresult{red and bold}, and the second-best results are
\secondresult{blue and underlined}.}
\label{tab:numerical_full}
\resizebox{\textwidth}{!}{%
\begin{tabular}{cc|cc|cc|cc|cc|cc|cc|cc|cc|cc|cc|cc}
\toprule
\multicolumn{2}{c}{\raisebox{1.1ex}{Models}}
& \multicolumn{2}{c}{\shortstack{\textbf{WorldTS}\\(ours)}}
& \multicolumn{2}{c}{\shortstack{KITE\\(2026)}}
& \multicolumn{2}{c}{\shortstack{DAG\\(2026)}}
& \multicolumn{2}{c}{\shortstack{GCGNet\\(2026)}}
& \multicolumn{2}{c}{\shortstack{DUET\\(2025)}}
& \multicolumn{2}{c}{\shortstack{CrossLinear\\(2025)}}
& \multicolumn{2}{c}{\shortstack{Amplifier\\(2025)}}
& \multicolumn{2}{c}{\shortstack{TimeKAN\\(2025)}}
& \multicolumn{2}{c}{\shortstack{TimeXer\\(2024)}}
& \multicolumn{2}{c}{\shortstack{TiDE\\(2023)}}
& \multicolumn{2}{c}{\shortstack{TFT\\(2021)}} \\
\multicolumn{2}{c}{Metrics}
& \multicolumn{1}{c}{mse} & \multicolumn{1}{c}{mae}
& \multicolumn{1}{c}{mse} & \multicolumn{1}{c}{mae}
& \multicolumn{1}{c}{mse} & \multicolumn{1}{c}{mae}
& \multicolumn{1}{c}{mse} & \multicolumn{1}{c}{mae}
& \multicolumn{1}{c}{mse} & \multicolumn{1}{c}{mae}
& \multicolumn{1}{c}{mse} & \multicolumn{1}{c}{mae}
& \multicolumn{1}{c}{mse} & \multicolumn{1}{c}{mae}
& \multicolumn{1}{c}{mse} & \multicolumn{1}{c}{mae}
& \multicolumn{1}{c}{mse} & \multicolumn{1}{c}{mae}
& \multicolumn{1}{c}{mse} & \multicolumn{1}{c}{mae}
& \multicolumn{1}{c}{mse} & \multicolumn{1}{c}{mae} \\
\midrule
\multirow[c]{3}{*}{\rotatebox{90}{NP}}
& 24 & \bestresult{0.171} & \bestresult{0.217} & \secondresult{0.179} & \secondresult{0.223}
& 0.202 & 0.237 & 0.208 & 0.237 & 0.246 & 0.287 & 0.210 & 0.266
& 0.252 & 0.303 & 0.273 & 0.310 & 0.236 & 0.266 & 0.284 & 0.301
& 0.219 & 0.249 \\
& 360 & \bestresult{0.423} & \bestresult{0.420} & \secondresult{0.471} & \secondresult{0.422}
& 0.521 & 0.451 & 0.531 & 0.459 & 0.576 & 0.528 & 0.531 & 0.508
& 0.587 & 0.534 & 0.538 & 0.529 & 0.600 & 0.475 & 0.601 & 0.498
& 0.539 & 0.501 \\
\cmidrule(lr){2-24}
& Avg
& \bestresult{0.297} & \bestresult{0.319} & \secondresult{0.325} & \secondresult{0.323}
& 0.362 & 0.344 & 0.370 & 0.348 & 0.411 & 0.407 & 0.371 & 0.387
& 0.420 & 0.418 & 0.405 & 0.419 & 0.418 & 0.371 & 0.443 & 0.400
& 0.379 & 0.375 \\
\midrule
\multirow[c]{3}{*}{\rotatebox{90}{PJM}}
& 24 & \bestresult{0.056} & \bestresult{0.141} & \bestresult{0.056} & \secondresult{0.143}
& \secondresult{0.057} & \secondresult{0.143} & 0.060 & 0.150 & 0.072 & 0.166
& 0.088 & 0.191 & 0.096 & 0.208 & 0.115 & 0.244 & 0.075 & 0.166
& 0.106 & 0.214 & 0.095 & 0.195 \\
& 360 & \bestresult{0.118} & \secondresult{0.216} & 0.135 & \bestresult{0.214}
& 0.130 & 0.218 & \secondresult{0.129} & 0.223 & 0.131 & 0.228
& 0.135 & 0.254 & 0.177 & 0.285 & 0.162 & 0.281 & 0.140 & 0.231
& 0.177 & 0.279 & 0.133 & 0.219 \\
\cmidrule(lr){2-24}
& Avg
& \bestresult{0.087} & \bestresult{0.178} & 0.096 & \bestresult{0.178}
& \secondresult{0.093} & \secondresult{0.180} & 0.095 & 0.187 & 0.102 & 0.197 & 0.112 & 0.223
& 0.137 & 0.246 & 0.139 & 0.262 & 0.108 & 0.198 & 0.142 & 0.246
& 0.114 & 0.207 \\
\midrule
\multirow[c]{3}{*}{\rotatebox{90}{BE}}
& 24 & \bestresult{0.340} & \bestresult{0.226} & \secondresult{0.348} & 0.240
& 0.361 & \secondresult{0.229} & 0.350 & 0.248 & 0.432 & 0.272 & 0.391 & 0.259
& 0.471 & 0.339 & 0.451 & 0.319 & 0.392 & 0.253 & 0.426 & 0.285
& 0.426 & 0.272 \\
& 360 & \bestresult{0.447} & \bestresult{0.293} & 0.507 & 0.332
& 0.485 & 0.330 & 0.511 & 0.340 & 0.597 & 0.436 & 0.568 & 0.416
& 0.646 & 0.487 & 0.645 & 0.495 & 0.512 & 0.327 & 0.571 & 0.364
& \secondresult{0.482} & \secondresult{0.310} \\
\cmidrule(lr){2-24}
& Avg
& \bestresult{0.393} & \bestresult{0.260} & 0.427 & 0.286
& \secondresult{0.423} & \secondresult{0.279} & 0.431 & 0.294 & 0.514 & 0.354
& 0.479 & 0.337 & 0.559 & 0.413 & 0.548 & 0.407 & 0.452 & 0.290
& 0.498 & 0.325 & 0.454 & 0.291 \\
\midrule
\multirow[c]{3}{*}{\rotatebox{90}{FR}}
& 24 & \secondresult{0.337} & \secondresult{0.176} & \bestresult{0.311} & 0.178
& 0.355 & \bestresult{0.171} & 0.347 & 0.188 & 0.384 & 0.251 & 0.390 & 0.226
& 0.459 & 0.348 & 0.454 & 0.296 & 0.366 & 0.208 & 0.418 & 0.255
& 0.543 & 0.253 \\
& 360 & \bestresult{0.431} & \bestresult{0.244} & \secondresult{0.462} & 0.271
& 0.473 & 0.268 & 0.482 & 0.279 & 0.607 & 0.403 & 0.575 & 0.370
& 0.648 & 0.468 & 0.641 & 0.452 & 0.489 & 0.273 & 0.551 & 0.308
& 0.465 & \secondresult{0.261} \\
\cmidrule(lr){2-24}
& Avg
& \bestresult{0.384} & \bestresult{0.210} & \secondresult{0.387} & 0.225
& 0.414 & \secondresult{0.219} & 0.415 & 0.234 & 0.495 & 0.327 & 0.482 & 0.298
& 0.554 & 0.408 & 0.547 & 0.374 & 0.427 & 0.241 & 0.484 & 0.281
& 0.504 & 0.257 \\
\midrule
\multirow[c]{3}{*}{\rotatebox{90}{DE}}
& 24 & \secondresult{0.274} & \secondresult{0.323} & \bestresult{0.262} & 0.327
& 0.277 & \bestresult{0.322} & 0.280 & 0.331 & 0.376 & 0.378 & 0.387 & 0.396
& 0.394 & 0.407 & 0.399 & 0.412 & 0.339 & 0.362 & 0.367 & 0.383
& 0.380 & 0.383 \\
& 360 & \bestresult{0.407} & \bestresult{0.382} & \secondresult{0.437} & \secondresult{0.413}
& 0.462 & 0.418 & 0.523 & 0.447 & 0.589 & 0.482 & 0.583 & 0.507
& 0.551 & 0.474 & 0.547 & 0.479 & 0.610 & 0.474 & 0.630 & 0.511
& 0.599 & 0.509 \\
\cmidrule(lr){2-24}
& Avg
& \bestresult{0.340} & \bestresult{0.353} & \secondresult{0.350} & \secondresult{0.370}
& 0.370 & \secondresult{0.370} & 0.401 & 0.389 & 0.482 & 0.430 & 0.485 & 0.452
& 0.473 & 0.441 & 0.473 & 0.445 & 0.475 & 0.418 & 0.499 & 0.447
& 0.489 & 0.446 \\
\midrule
\multirow[c]{3}{*}{\rotatebox{90}{Energy}}
& 24 & \bestresult{0.039} & \bestresult{0.151} & \secondresult{0.071} & \secondresult{0.208}
& 0.079 & 0.215 & 0.081 & 0.221 & 0.117 & 0.283 & 0.241 & 0.418
& 0.138 & 0.306 & 0.135 & 0.298 & 0.122 & 0.273 & 0.103 & 0.248
& 0.093 & 0.235 \\
& 360 & \bestresult{0.142} & \bestresult{0.300} & \secondresult{0.151} & \secondresult{0.305}
& 0.169 & 0.320 & 0.182 & 0.334 & 0.288 & 0.452 & 0.237 & 0.385
& 0.328 & 0.472 & 0.302 & 0.464 & 0.204 & 0.357 & 0.202 & 0.355
& 0.167 & 0.331 \\
\cmidrule(lr){2-24}
& Avg
& \bestresult{0.091} & \bestresult{0.226} & \secondresult{0.111} & \secondresult{0.257}
& 0.124 & 0.267 & 0.131 & 0.277 & 0.203 & 0.367 & 0.239 & 0.401
& 0.233 & 0.389 & 0.218 & 0.381 & 0.163 & 0.315 & 0.153 & 0.302
& 0.130 & 0.283 \\
\midrule
\multirow[c]{3}{*}{\rotatebox{90}{Sdwpfm1}}
& 24 & 0.384 & \bestresult{0.399} & 0.384 & 0.410
& \bestresult{0.351} & \secondresult{0.400} & 0.376 & 0.415 & 0.551 & 0.564
& \secondresult{0.355} & 0.473 & 0.364 & 0.445 & 0.418 & 0.503 & 0.558 & 0.533
& 0.474 & 0.488 & 0.366 & 0.421 \\
& 360 & \bestresult{0.398} & \bestresult{0.425} & \secondresult{0.450} & \secondresult{0.491}
& 0.495 & 0.522 & 0.472 & 0.499 & 0.646 & 0.577 & 0.497 & 0.532
& 0.510 & 0.535 & 0.476 & 0.566 & 0.845 & 0.684 & 0.492 & 0.526
& 0.597 & 0.528 \\
\cmidrule(lr){2-24}
& Avg
& \bestresult{0.391} & \bestresult{0.412} & \secondresult{0.417} & \secondresult{0.451}
& 0.423 & 0.461 & 0.424 & 0.457 & 0.599 & 0.570 & 0.426 & 0.502
& 0.437 & 0.490 & 0.447 & 0.534 & 0.701 & 0.609 & 0.483 & 0.507
& 0.482 & 0.474 \\
\midrule
\multirow[c]{3}{*}{\rotatebox{90}{Sdwpfm2}}
& 24 & 0.417 & \bestresult{0.414} & 0.443 & 0.450
& \bestresult{0.372} & \bestresult{0.414} & 0.421 & \secondresult{0.441} & 0.445 & 0.452
& 0.477 & 0.536 & \secondresult{0.394} & 0.462 & 0.474 & 0.538 & 0.627 & 0.570
& 0.461 & 0.492 & 0.411 & 0.458 \\
& 360 & \bestresult{0.474} & \bestresult{0.486} & \secondresult{0.506} & \secondresult{0.515}
& 0.583 & 0.556 & 0.529 & 0.531 & 0.584 & 0.528 & 0.589 & 0.611
& 0.587 & 0.563 & 0.520 & 0.589 & 0.978 & 0.736 & 0.511 & 0.540
& 0.541 & 0.519 \\
\cmidrule(lr){2-24}
& Avg
& \bestresult{0.446} & \bestresult{0.450} & \secondresult{0.475} & \secondresult{0.483}
& 0.477 & 0.485 & \secondresult{0.475} & 0.486 & 0.514 & 0.490 & 0.533 & 0.573
& 0.491 & 0.512 & 0.497 & 0.564 & 0.803 & 0.653 & 0.486 & 0.516
& 0.476 & 0.488 \\
\midrule
\multirow[c]{3}{*}{\rotatebox{90}{Sdwpfh1}}
& 24 & \bestresult{0.394} & \bestresult{0.435} & 0.424 & 0.464
& 0.408 & \secondresult{0.438} & 0.435 & 0.486 & 0.527 & 0.513
& 0.548 & 0.585 & 0.576 & 0.627 & 0.511 & 0.582 & 0.651 & 0.587
& 0.434 & 0.489 & \secondresult{0.401} & 0.460 \\
& 360 & \bestresult{0.357} & \bestresult{0.428} & \secondresult{0.457} & 0.517
& 0.489 & 0.534 & 0.465 & \secondresult{0.514} & 0.551 & 0.519
& 0.566 & 0.601 & 0.497 & 0.569 & 0.643 & 0.694 & 0.841 & 0.700
& 0.472 & 0.527 & 0.557 & 0.523 \\
\cmidrule(lr){2-24}
& Avg
& \bestresult{0.376} & \bestresult{0.431} & \secondresult{0.441} & 0.491
& 0.448 & \secondresult{0.486} & 0.450 & 0.500 & 0.539 & 0.516 & 0.557 & 0.593
& 0.537 & 0.598 & 0.577 & 0.638 & 0.746 & 0.643 & 0.453 & 0.508
& 0.479 & 0.491 \\
\midrule
\multirow[c]{3}{*}{\rotatebox{90}{Sdwpfh2}}
& 24 & \bestresult{0.406} & \bestresult{0.451} & 0.477 & 0.498
& \secondresult{0.438} & \secondresult{0.465} & 0.473 & 0.506 & 0.629 & 0.563
& 0.468 & 0.540 & 0.473 & 0.533 & 0.580 & 0.614 & 0.820 & 0.677
& 0.579 & 0.553 & 0.474 & 0.493 \\
& 360 & \bestresult{0.432} & \bestresult{0.470} & \secondresult{0.524} & 0.555
& 0.608 & 0.595 & 0.566 & 0.565 & 0.665 & 0.569 & 0.608 & 0.609
& 0.569 & 0.628 & 0.713 & 0.729 & 0.962 & 0.761 & 0.619 & 0.614
& 0.657 & \secondresult{0.549} \\
\cmidrule(lr){2-24}
& Avg
& \bestresult{0.419} & \bestresult{0.461} & \secondresult{0.500} & 0.526
& 0.523 & 0.530 & 0.520 & 0.536 & 0.647 & 0.566 & 0.538 & 0.574
& 0.521 & 0.581 & 0.647 & 0.672 & 0.891 & 0.719 & 0.599 & 0.583
& 0.566 & \secondresult{0.521} \\
\midrule
\multirow[c]{3}{*}{\rotatebox{90}{Colbun}}
& 10 & \bestresult{0.052} & \bestresult{0.080} & \secondresult{0.055} & 0.101
& 0.061 & \secondresult{0.094} & 0.065 & 0.108 & 0.089 & 0.134 & 0.071 & 0.102
& 0.071 & 0.121 & 0.061 & 0.101 & 0.113 & 0.172 & 0.089 & 0.131
& 0.092 & 0.135 \\
& 30 & \secondresult{0.125} & \bestresult{0.213} & \bestresult{0.121} & 0.243
& 0.135 & \secondresult{0.215} & 0.149 & 0.243 & 0.307 & 0.397 & 0.182 & 0.288
& 0.275 & 0.370 & 0.195 & 0.249 & 0.176 & 0.299 & 0.240 & 0.322
& 0.383 & 0.460 \\
\cmidrule(lr){2-24}
& Avg
& \bestresult{0.088} & \bestresult{0.146} & \bestresult{0.088} & 0.172
& \secondresult{0.098} & \secondresult{0.154} & 0.107 & 0.175 & 0.198 & 0.266 & 0.126 & 0.195
& 0.173 & 0.246 & 0.128 & 0.175 & 0.145 & 0.235 & 0.164 & 0.227
& 0.238 & 0.297 \\
\midrule
\multirow[c]{3}{*}{\rotatebox{90}{Rapel}}
& 10 & 0.170 & \bestresult{0.200} & 0.192 & 0.231
& \bestresult{0.151} & \secondresult{0.203} & 0.211 & 0.230 & 0.174 & 0.219
& \secondresult{0.163} & 0.209 & 0.181 & 0.227 & 0.174 & 0.231 & 0.301 & 0.308
& 0.228 & 0.271 & 0.201 & 0.253 \\
& 30 & \bestresult{0.286} & \bestresult{0.317} & \secondresult{0.296} & \secondresult{0.358}
& 0.309 & 0.408 & 0.401 & 0.384 & 0.365 & 0.432 & 0.340 & 0.417
& 0.333 & 0.416 & 0.325 & 0.390 & 0.387 & 0.416 & 0.411 & 0.432
& 0.409 & 0.414 \\
\cmidrule(lr){2-24}
& Avg
& \bestresult{0.228} & \bestresult{0.259} & 0.244 & \secondresult{0.294}
& \secondresult{0.230} & 0.305 & 0.306 & 0.307 & 0.269 & 0.326 & 0.252 & 0.313
& 0.257 & 0.321 & 0.249 & 0.311 & 0.344 & 0.362 & 0.320 & 0.351
& 0.305 & 0.333 \\
\bottomrule
\end{tabular}%
}
\end{table*}

\section{Limitations and Future Work}
\label{app:limitations}

Despite encouraging empirical results, a rigorous theoretical understanding
of covariate-conditioned predictive-state learning remains open. Future work
could formalize the links among multimodal covariate information,
predictive-state quality, and forecasting error.

WorldTS is an initial attempt to adapt latent-space predictive learning from
world models to time-series forecasting. Its objectives and regularization
largely follow general-purpose designs; regularization tailored to temporal
structure could better capture temporal dependencies and dynamics.

WorldTS currently targets regularly sampled series and is trained separately
for each dataset. Integrating this approach with pretrained time-series
foundation models and extending it to irregular sampling and asynchronous
covariates could improve transferability and
applicability.

\clearpage
\section{Extended Related Work}
\label{app:extended_related_work}

\subsection{Time-Series Forecasting}

Early time-series forecasting primarily relied on statistical models and classical machine learning.
With advances in deep learning, neural networks have demonstrated strong representation learning capabilities in visual generation and editing~\citep{ma2024followpose,ma2024followyouremoji,ma2025controllable,ma2026livelight} and have been widely applied across time-series applications~\citep{cheng2026star}.
Work on deep forecasting models has explored ways to improve forecasting performance through better representation learning and training objectives~\citep{wu2025srsnet,qiu2025DBLoss}, as well as probabilistic forecasting~\citep{wu2025k2vae}, cross-variable correlation modeling~\citep{cheng2026ccd}, and correlation-aware adaptation of foundation models~\citep{cheng2026cora}, extending to forecasting with irregular sampling~\citep{liu2026rethinking,liu2026astgi,qiu2026bridging} and missing observations~\citep{11002729,yu2025merlin}.

\subsection{Latent-Space Predictive Modeling}

Latent-space predictive models encode observations into states and model their future evolution in representation space.
DreamerV3 learns latent dynamics for decision-making through imagined trajectories~\citep{hafner2023dreamerv3}.
Joint-embedding approaches use encoded observations as prediction targets: I-JEPA predicts image-region representations~\citep{assran2023ijepa}, while V-JEPA~2-AC and LeWorldModel model action-conditioned future states from visual observations~\citep{assran2025vjepa2,maes2026leworldmodel}.
For time series, TimeAlign aligns historical and future representations~\citep{hu2026timealign}, and LatentTSF predicts future states in a shared latent space before decoding observations~\citep{yang2026latentsf}.
WorldTS incorporates multimodal covariates into directly supervised future-state prediction and uses two-stage training to separate learning state dynamics from decoding future observations.

\subsection{Covariate-aware Forecasting}

Covariate-aware forecasting uses external information to complement the target history.
TimeXer and CrossLinear model dependencies between target and exogenous series~\citep{wang2024timexer,zhou2025crosslinear}, while TFT and TiDE accommodate known future covariates~\citep{lim2021tft,das2023tide}.
DAG captures temporal and channel correlations~\citep{qiu2026dag}, GCGNet introduces graph-structured consistency~\citep{li2026gcgnet}, and KITE uses external information for knowledge-guided probabilistic forecasting~\citep{cheng2026kite}.
Multimodal approaches further exploit text and images: TaTS and VoT model textual context through temporal narratives and event-driven reasoning~\citep{li2026tats,wang2026vot}, FusionSF combines heterogeneous inputs for solar power forecasting~\citep{ma2024fusionsf}, and UniCA adapts foundation models to heterogeneous covariates~\citep{han2026unica}.
WorldTS focuses on using these covariates to condition latent future-state prediction, with representations of ground-truth future observations providing direct supervision.
This latent-space formulation provides a natural way to integrate heterogeneous multimodal information into future-state modeling.

\end{document}